\pdfoutput=1
\documentclass[11pt]{article}

\usepackage[final]{acl}

\usepackage{times}
\usepackage{latexsym}

\usepackage[T1]{fontenc}
\usepackage[utf8]{inputenc}

\usepackage{microtype}

\usepackage{inconsolata}

\usepackage{graphicx}
\usepackage{booktabs}
\usepackage{multirow}
\usepackage{booktabs}
\usepackage{hyperref}
\usepackage{enumitem}

\usepackage[most]{tcolorbox}
\definecolor{Magenta}{rgb}{0.8, 0.1, 0.6}

\title{What Makes a Good Reasoning Chain? \\ Uncovering Structural Patterns in Long Chain-of-Thought Reasoning}

\author{
  Gangwei Jiang\textsuperscript{1,2}\thanks{\hspace{1mm} Equal contribution.},
  Yahui Liu\textsuperscript{3}\footnotemark[1],
  Zhaoyi Li\textsuperscript{1,2},
  Qi Wang\textsuperscript{3},
  Fuzheng Zhang\textsuperscript{3}, \\
  \textbf{Linqi Song\textsuperscript{2}},
  \textbf{Ying Wei\textsuperscript{4}},
  \textbf{Defu Lian\textsuperscript{1}}
  \thanks{\hspace{1mm} Corresponding author.} \\  % Add or remove authors as needed
  \textsuperscript{1}University of Science and Technology of China ,
  \textsuperscript{2}City University of Hong Kong , \\
  \textsuperscript{3}Kuaishou Technology,
  \textsuperscript{4}Zhejiang University,  \\ % Add or remove affiliations as needed
  \texttt{gwjiang@mail.ustc.edu.cn} \\
}

\begin{document}
\maketitle
\begin{abstract}

Recent advances in reasoning with large language models (LLMs) have popularized Long Chain-of-Thought (LCoT), a strategy that encourages deliberate and step-by-step reasoning before producing a final answer. 
While LCoTs have enabled expert-level performance in complex tasks, %such as mathematics and scientific problem-solving, 
how the internal structures of their reasoning chains drive, or even predict, the correctness of final answers remains a critical yet underexplored question.
% assessing the quality of these extended reasoning chains remains an unresolved challenge. 
% Prior methods, particularly process reward models (PRMs), rely heavily on step-wise semantic evaluation and fail to scale effectively with longer, more complex CoTs. 
In this work, we present LCoT2Tree, an automated framework that converts sequential LCoTs into hierarchical tree structures and thus enables deeper structural analysis of LLM reasoning. 
Using graph neural networks (GNNs), we reveal that structural patterns extracted by LCoT2Tree, including exploration, backtracking, and verification, serve as stronger predictors of final performance across a wide range of tasks and models.
% Our method supports structure-aware analysis and enables quantitative evaluation using graph neural networks (GNNs), leading to more accurate assessments of reasoning quality across a wide range of tasks and models. 
Leveraging an explainability technique, we further identify critical thought patterns such as over-branching that account for failures. Beyond diagnostic insights, the structural patterns by LCoT2Tree support practical applications, including improving Best-of-N decoding effectiveness. Overall, our results underscore the critical role of internal structures of reasoning chains, positioning LCoT2Tree as a powerful tool for diagnosing, interpreting, and improving reasoning in LLMs.

\end{abstract}

%----------------------%
\section{Introduction}
\label{sec:introduction}
%----------------------%
Large Language Models (LLMs) have achieved remarkable progress in nature language understanding and processing, with recent developments extending their capabilities to more complex reasoning tasks. Cutting-edge models such as OpenAI o3~\citep{o3-mini} and DeepSeek R1~\citep{guo2025deepseek} push this frontier by emulating System 2 thinking~\citep{li2025system}, \emph{i.e}, engaging in slow, deliberate, and step-by-step reasoning before arriving at a final answer. This approach, %structured reasoning is commonly
well known as Long Chain-of-thought (LCoT) reasoning~\citep{chen2025towards, gandhi2025cognitive}, has empowered LLMs to achieve expert-level performance in challenging tasks such as mathematics, code generation, and scientific problem-solving~\citep{seed2025seed, team2025kimi, teamqwq}. 
% However, despite their growing adoption, the specific forms or structures of LCoT that most effectively enhance reasoning quality remain poorly understood.
% However, despite 
Despite their growing adoption, 
LCoTs remain largely a black box in one key aspect: \emph{what makes a good thought chain?}
%how their internal structures govern the success or failure of reasoning}?

% effectively assessing the quality of these reasoning chains remains an open challenge.
% how different LCoT components influence the reasoning quality remain poorly understood.

\begin{figure}[t]
  \includegraphics[width=\columnwidth]{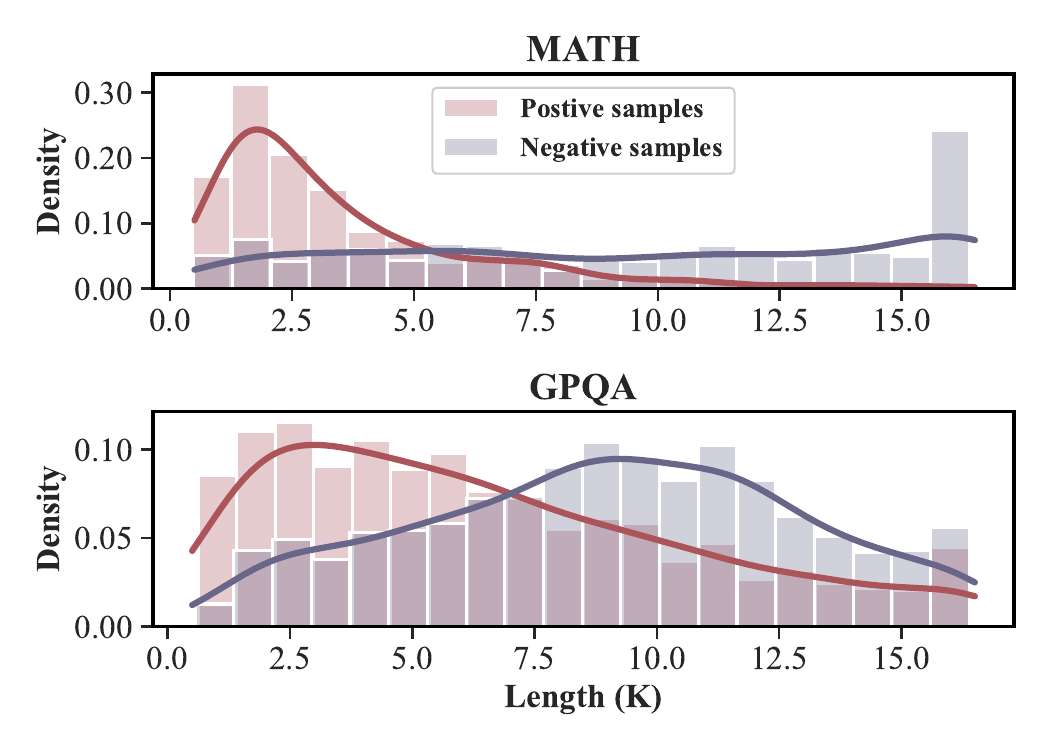}
  \vspace{-2em}
  \caption{The distribution of output token length for correctly answered (Positive) and incorrectly answered (Negative) samples by DeepSeek-R1-Distill-Qwen-32B on two datasets.}
  \vspace{-1em}
  \label{fig:sec3.1:distrib}
\end{figure}

Before the emergence of LCoT, researchers attempted to answer this question from a semantic perspective, often using process reward models (PRMs) that provide token-level or step-wise supervision based on logical coherence and factual accuracy~\citep{xia2025evaluating, zhang2025lessons}. 
While effective for short or moderately long CoTs, PRMs struggle to scale effectively as the length and structure complexity of reasoning chains increase~\citep{he2025can}.  
In the LCoT era, recent work has increasingly emphasized the importance of reasoning structure~\citep{gandhi2025cognitive, li2025llms,ye2025limo}. Both \citeauthor{wu2025more} and \citeauthor{, ballon2025relationship} highlighted the overthinking phenomenon, where overly long reasoning chains can degrade rather than improve final answer quality.
%(Figure~\ref{fig:sec3.1:motivation})\yahui{Fig 2 appears before Fig 1}. 
However, our analysis (Figure~\ref{fig:sec3.1:distrib}) shows that response length alone remains an inadequate predictor of answer correctness, as responses with similar lengths vary greatly
in correctness.
These findings suggest that these heuristics, such as token count, step count, or PRM-based semantic metrics fall short in effectively dictating reasoning success.
%evaluating the quality of reasoning in LLMs.

Thus, we  
%To address this, we 
propose Long Chain-of-Thought to Tree (LCoT2Tree), the first automated framework for structural analysis of reasoning in LLMs.
LCoT2Tree transforms sequential LCoTs into hierarchical tree representations (Section~\ref{sec3.2}), which enable 
 structural patterns in reasoning chains, including
% By uncovering the underlying cognitive behaviors, such as 
exploration, backtracking, and verification, to be made explicit and analyzable. %LCoT2Tree enables a richer, structure-aware perspective on model reasoning. 
By modeling these trees with graph neural networks (GNNs), we not only extract these structural patterns as features, but also demonstrate that they serve as strong predictors of reasoning success %which greatly improve the 
% evaluate reasoning quality and 
% differentiation between successful and flawed reasoning 
(Section~\ref{sec3.3}).
% Our approach significantly improves the ability to distinguish between correct and incorrect reasoning, advancing beyond token-level heuristics and offering a principled path toward structure-aware evaluation

Beyond establishing their predictive power, we further investigate which structural patterns specifically contribute to reasoning success or failure, how these patterns vary across tasks and models, and how they can be applied to further enhance LLM reasoning in practice. 
% Our framework offers both diagnostic insights into reasoning behaviors and practical benefits. 
Concretely, by leveraging a GNN-based explainability method, we unveil key thought patterns (\textit{i.e.}, critical substructures within the tree) that that explain answer correctness across diverse tasks and models
%contribute to different reasoning behaviors 
(Section~\ref{sec4}). These analyses reveal how reasoning behaviors differ by (1) answer correctness, (2) task type, and (3) model variant.
% highlighting the value of structural information in reasoning analysis.
% Beyond behavior analysis, 
Furthermore, we demonstrate that these patterns can be leveraged to improve %LCoT2Tree also supports practical applications such as 
Best-of-N decoding:  %Specifically, we show that
incorporating our tree-based predictive classifier into its selection strategy consistently enhances accuracy across diverse models and tasks (Section~\ref{sec5}).
% This representation supports more accurate reasoning evaluation and opens the door to new applications in behavior diagnosis, decoding optimization, and robustness analysis.
We summarize the main contributions of the proposed LCoT2Tree in three aspects:
\begin{itemize}[leftmargin=*, nolistsep, noitemsep]
    \item (\emph{Predictability}) We are the first to explicitly construct structural representations of LCoT; our proposed LCoT2Tree offers stronger signals for reasoning success and improves binary classification of answer correctness by an average of \textbf{5.63\%},
    compared to using length alone.
    % by modeling the underlying reasoning structure.
    % (Section~\ref{sec3})
    \item (\emph{Interpretability}) We leverage LCoT2Tree to pinpoint the reasoning patterns that oftentimes lead to errors, \emph{e.g.}, over-branching, and to account for disparate behaviors across tasks and models.
    %, revealing erroneous patterns such as over-branching and qualitative differences across settings. 
    % (Section~\ref{sec4})
    \item (\emph{Practicality}) We demonstrate that LCoT2Tree offers a principled path for selecting well-structured reasoning chains, greatly enhancing Best-of-N decoding and also remaining extensible for future decoding strategies.

    % offers a strong signal for evaluating reasoning quality in Best-of-N decoding, enhancing the selection of structurally sound outputs. 
    % (Section~\ref{sec5})
    % Reasoning structure plays a crucial role in evaluating reasoning quality. 
    % Interestingly, we find that the prediction of answer correctness for models can be enhanced by the structure representation provided by our LCoT2Tree tool. 
    % (Section~\ref{sec3})
    % \item We are the first to visualize the error patterns in the LCoT. It demonstrates that over-branching during a specific reasoning step not only needlessly complicates the reasoning process but also significantly increases the probability of errors. (Section~\ref{subsec:4.1_error_patterns})
    % \item Long CoT varies significantly across tasks and models, and these differences can be reliably identified. Moreover, our proposed tree-based classification method demonstrates strong effectiveness in capturing such distinctions. (Section~\ref{subsec:4.2_task_specific} and \ref{subsec:4.3_model_specific})
    % \item 
    % LCoT2Tree has been verified to serve as an effective process feedback in Best-of-N (BoN) decoding, evaluating whether the reasoning steps are reasonable. (Section~\ref{sec5})
\end{itemize}

%\begin{itemize}
%    \item Reasoning structure plays a crucial role in evaluating reasoning quality. Interestingly, we find that the prediction of answer correctness for all models (\textit{e.g.}, Qwen, DeepSeek, Doubao and Gemini families) can be enhanced by the structure representation provided by our LCoT2Tree tool. (Section~\ref{sec3})
%    \item We are the first to visualize the error patterns in the LCoT. It demonstrates that over-branching during a specific reasoning step not only needlessly complicates the reasoning process but also significantly increases the probability of errors. (Section~\ref{subsec:4.1_error_patterns})
%    \item Long CoT shows obvious differences in different tasks, and such differences can be identified. Moreover, our proposed graph-based classification method proves to be highly effective. (Section~\ref{subsec:4.2_task_specific})
%    \item Owing to relatively small differences in the training strategies, data, and other aspects of the reasoning models pose a challenge to differentiate the reasoning structures between their LCoTs. (Section~\ref{subsec:4.3_model_specific})
%\end{itemize}
%------------------------%
\section{Related Works}
\label{sec:related_work}
%------------------------%

\paragraph{Reasoning LLMs}
Advancing reasoning capabilities of LLMs has shown benefits in tackling complex tasks~\citep{kojima2022large,wei2022chain,li2025system}. 
%and increasing their intelligence~\citep{li2025system}. 
Researchers first demonstrated that %generating 
%intermediate logical steps before reaching a final answer, called
CoT prompting 
can significantly improve performance on complex tasks like arithmetic~\citep{wei2022chain}. 
%Subsequently, to 
To refine the reasoning processes, hierarchical cognitive phases have been introduced, such as multi-path exploration~\citep{wang2023selfconsistency, zhou2023leasttomost, yao2023tree}, step verification~\citep{miao2024selfcheck, gou2024critic}, and iterative refinement~\citep{madaan2023self, besta2024graph}. These approaches expand solution spaces and deepen reasoning, driving more reliable answers. More recently, models such as Deepseek-R1~\citep{guo2025deepseek}, Kimi-1.5~\citep{team2025kimi} and QwQ-32B~\citep{teamqwq} have leveraged rule-based reinforcement learning to embed reasoning capabilities directly into model parameters, %introducing long CoT reasoning. 
achieving remarkable progress in handling complex tasks~\cite{chen2025towards, gandhi2025cognitive}.
%These models driven remarkable progress excel in handling tasks through thinking with extensive exploration, rigorous reflection, and deep reasoning~\citep{chen2025towards, gandhi2025cognitive}. 

\paragraph{Chain-of-Thought Analysis}
Numerous studies have explored when CoT prompting is effective. Empirical research has revealed that factors, such as step length~\citep{jin2024impact}, relevance, the order of reasoning fragments~\citep{wang2023towards}, and prompt structure~\citep{li2025llms}, heavily influence performance. Expanding on these findings, ~\citet{feng2023towards} and  ~\citet{chen2024unlocking} proposed that there is an inherent reasoning limit in LLMs when tackling tasks exceeding a complexity threshold.
% Additionally, they study the CoT reasoning from a mechanism interpretability perspective, concluding that LLMs deploy multiple parallel answer generation paths internally (Yang et al., 2024; Li et al., 2024a; Dutta 057 et al., 2024)
%
In the context of long CoT, research has increasingly emphasized the importance of response structures in enhancing reasoning success~\citep{li2025llms, gandhi2025cognitive, muennighoff2025s1, ye2025limo}.  Additionally, challenges like the overthinking phenomenon, %persist, 
where overly long responses inadvertently hurt the model performance, seemingly establish the correlation between length and reasoning success~\citep{chen2024not, wu2025more, cuadron2025danger}. 
%While prior work has relied on manual analysis to study the influence of response components, 
% In contrast to
Beyond these
prior works, we are motivated to develop an automated tool to empirically identify the
%components and extract 
structural patterns that dictate reasoning success in long CoT.
%responses. 
%Using this tool, we reveal a strong correlation between response structure and final answer accuracy.
% , highlighting the critical role of structured reasoning in achieving optimal performance.

% \paragraph{Chain-of-Thought Evaluation}
%Evaluating the quality of generated reasoning chains remains challenging. 
Besides the above analyses towards reasoning success, another line of works primarily analyzes towards the semantic rationality of reasoning. Thus, early methods directly %relied on reference-based evaluations 
compare generated steps to human-annotated explanations~\citep{welleck2022naturalprover}.
However, such methods often fail to capture logical coherence beyond surface-level similarity. 
%In response, reference-free evaluation techniques have been developed, integrating tools~\citep{saparov2023language, gou2024critic} or rules~\citep{prasad2023receval, golovneva2023roscoe} to assess the rationality of intermediate reasoning steps. 
Recent work has introduced LLM-driven PRMs~\citep{ling2023deductive, yuan2024free,she2025r,zhang2025lessons} to provide holistic and step-wise assessment, but they
%which provides holistic and step-wise supervision to both isolate factual errors and quantify logical consistency.  Despite these advancements, PRMs 
struggle with scaling to long and complex CoT reasoning chains~\citep{he2025can}.  
%To address this, 
Rather than relying on surface similarity or token-level reward signals, we analyze reasoning success through internal structural patterns derived from hierarchical tree representations.
The structural patterns offer a principled alternative for dictating ``good'' chains, and are fully complementary to this body of semantic-based research.
% , established in our work, provide a structural assessment towards reasoning success, being complementary to semantic accuracy in this line of research.
%designed specifically 
% to evaluate long CoT quality, 
%This framework complements existing methods by 
% focusing on response structure rather than solely semantic accuracy.
% In contrast to prior works that prioritize semantic understanding, we introduce a structural assessment framework to evaluate long CoT efficiency. Our approach successfully serves as a valuable complement to existing PRM approaches.
%--------------------------%
\section{LCoT2Tree: Automated Long Chain-of-Thoughts to Tree }
\label{sec3}
%--------------------------%
% we introduce an automated tool for analyzing Long Chains-of-Thought (LCoT) generated by reasoning LLMs.
In this section, we empirically study overthinking, highlighting issues with assessing reasoning quality via CoT length. 
Then, we propose Long Chain-of-Thought to Tree (LCoT2Tree), an automated tool that converts LCoTs into tree structures to reveal cognitive frameworks and enable deeper analysis of LLMs' reasoning processes.
%This tool reveals the underlying cognitive framework, providing a clearer and more insightful way to analyze the reasoning processes of LLMs. 

%-----------------------------------%
\subsection{Overthinking Phenomenon}
\label{sec3.1}
\begin{figure}[t]
  \includegraphics[width=\columnwidth]{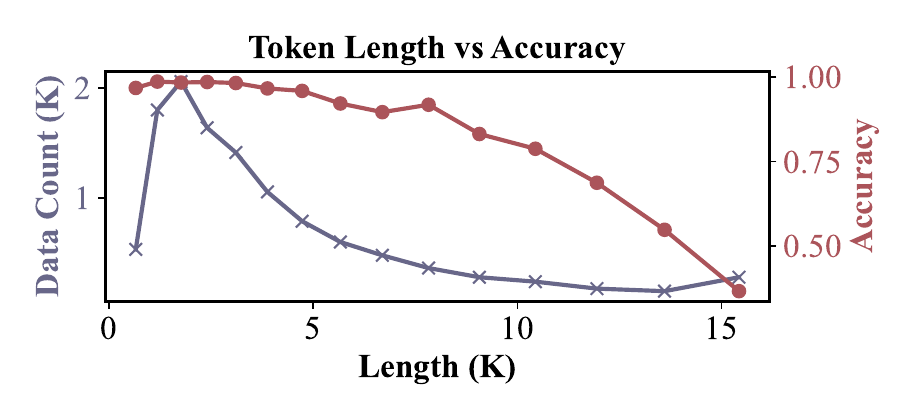}
  \vspace{-2em}
  \caption{Data count and accuracy of the MATH dataset for DeepSeek-32B across varying response lengths. Accuracy notably declines as response length increases.}
  \label{fig:sec3.1:motivation}
  \vspace{-1em}
\end{figure}

%-----------------------------------%
The ``overthinking'' phenomenon in reasoning models refers to situations where a model expends excessive computational resources (\textit{e.g.}, generating overly long sequences or repeating reasoning steps), yet makes little contributions to the correctness of final answer. In some cases, this can even lead to a decline in performance~\cite{chen2024not, wu2025more}. Figure~\ref{fig:sec3.1:motivation} illustrates this phenomenon by showing the relationship between the output token length and the answer accuracy of DeepSeek-32B (\textit{i.e.}, DeepSeek-R1-Distill-Qwen-32B~\cite{guo2025deepseek}) on the MATH~\citep{hendrycks2021measuring} dataset. It demonstrates that as the reasoning chain becomes unnecessarily long, model performance deteriorates, highlighting how overthinking can harm the reasoning ability of LLMs.

To tackle this issue, researchers have proposed using a length penalty during the training period to constrain the length of generated LCoTs
%, with the goal of promoting more concise and effective reasoning
~\citep{team2025kimi, yu2025dapo}. However, this strategy relies on the oversimplified assumption that shorter or moderately long reasoning chains inherently lead to better reasoning quality. In this work, we 
%challenge that assumption by conducting 
conduct
a classification experiment to empirically quantify the actual relationship between these two factors and uncover the limitations of relying on length as an indicator of reasoning quality.

\begin{figure*}[t]
  \includegraphics[width=\textwidth]{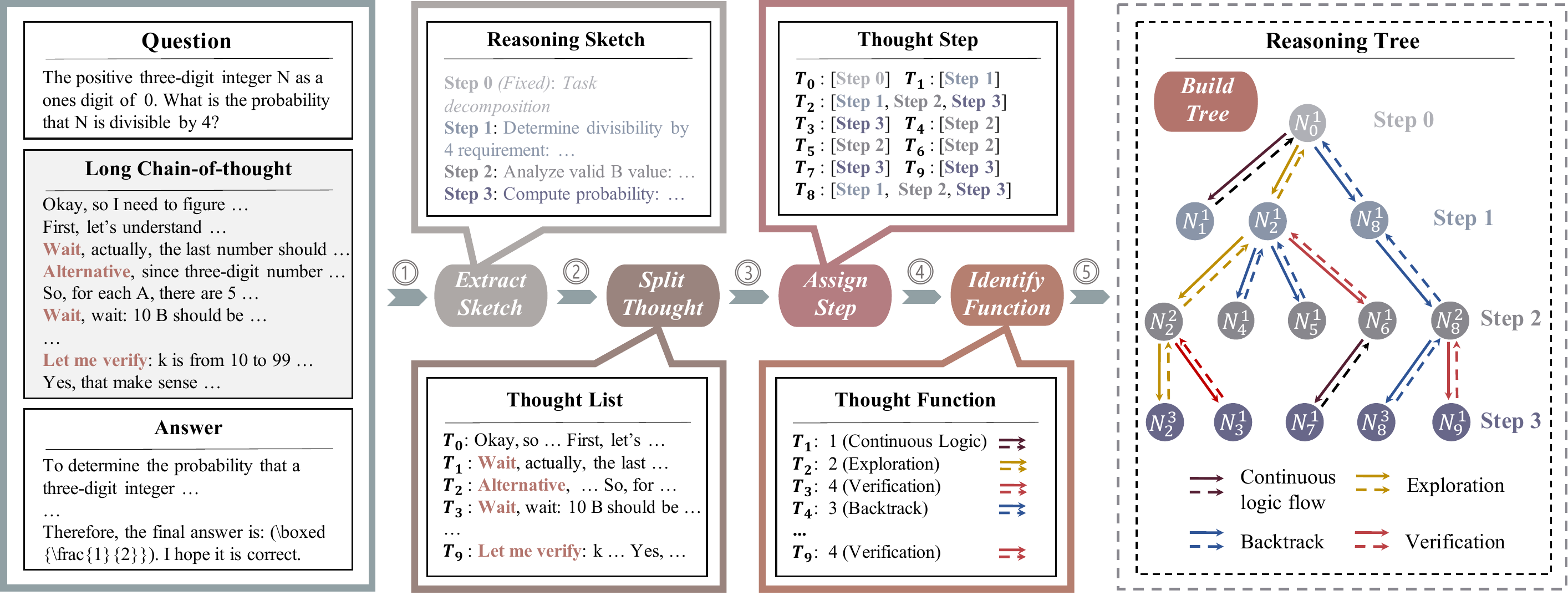}
  \vspace{-2em}
  \caption{The workflow for LCoT2Tree. It transforms sequential long chain-of-thought into reasoning tree, which involves five steps: (1) Extract Sketch, (2) Split Thought, (3) Assign Step, (4) Identify Function, and (5) Build Tree.}
  \vspace{-1em}
  \label{fig:sec3.2:lcot2tree}
\end{figure*}

\noindent \textbf{Experimental Setup.} For our study, we use 
DeepSeek-32B, DeepSeek-R1~\citep{guo2025deepseek}, QwQ-32B~\citep{teamqwq}, Seed-1.5-Thinking-pro~\citep{seed2025seed}, and Grok-3-mini-beta~\citep{grok} as the primary models. We evaluate these models on four benchmark datasets: MATH (Level5 question in high school math competitions;~\citealp{hendrycks2021measuring}), GPQA (``main'' subset in grade-level google-proof question answering;~\citealp{rein2024gpqa}), LiveCodeBench (version 5 in live code benchmark;~\citealp{jain2025livecodebench}), and MMLU-Pro (proficient-level multi-discipline language understanding;~\citealp{wang2024mmlu}). For each dataset, we collect 2,000 model responses, consisting of 1,000 correctly answered cases (Positive) and 1,000 incorrectly answered cases (Negative). These samples are divided into training and testing sets at a ratio of 4:1. 

In our experiments, we train a logistic regression model using LCoT response length as the input feature and answer correctness as the target label. The test set accuracy quantifies the degree of correlation between LCoT length and reasoning quality-higher accuracy suggests a stronger association between these two factors.

% \begin{table}[]
% \begin{center}
% \begin{small}
% \begin{tabular}{@{}l|cccc@{}}
% \toprule
%         & MATH    & GPQA    & LCB & MMLU-P \\ \midrule
% DS-32B  &  74.13\% & 67.08\% & 81.59\% & 59.95\% \\
% QwQ-32B & 75.82\% & 62.09\% & 78.30\% & 58.00\% \\ \bottomrule
% \end{tabular}
% \caption{Results of classifying the answer correctness according to the LCoT length. We test two reasoning models (\textit{i.e.}, DeepSeek-32B and QwQ-32B) on four public benchmarks.}
% \label{tab:sec3.1:length-corr}
% \end{small}
% \end{center}
% \end{table}

\noindent \textbf{Results and Analysis.} 
Figure~\ref{fig:sec3.1:distrib} shows the token length distributions of positive and negative samples. 
It reveals a significant overlap between the two classes, indicating that responses with similar lengths can vary greatly in reasoning quality. 
Moreover, Table~\ref{tab:sec3.3:main_compare} presents the classification results, where the accuracy on the MMLU-Pro dataset is only 60.0\% for DeepSeek-32B and 58.0\% for QwQ-32B. These relatively low accuracies underscore the limitations of using response length alone in predicting reasoning success. 
%To explore this further, we visualize the token length distributions of positive and negative samples in Figure~\ref{fig:sec3.1:distrib}. 
% (More cases in Appendix~\ref{appendix:sec3.1}). 
%The figure reveals a significant overlap between the two classes, indicating that responses with similar lengths can vary greatly in reasoning quality.  
% which again confirms that response length alone is not a reliable indicator of reasoning quality.

%-----------------------------------%
\subsection{LCoT2Tree Tool}
\label{sec3.2}
%-----------------------------------%
%Building on the limitations identified in the length-based approach to evaluating LCoT quality, this paper introduces a novel tool, LCoT2Tree. 
We present LCoT2Tree, a novel tool that extracts structural insights from LCoTs, addressing the limitations of length-based prediction.
% We introduce a novel tool, named LCoT2Tree, to overcome the limitations of length-based approaches for measuring LCoT quality.
%The tool enables fine-grained extraction of the structural information within LCoTs, offering deeper insights into how the different reasoning patterns influence overall effectiveness. 
LCoT2Tree converts the sequential chain of reasoning into a tree structure, enabling a deeper analysis of cognitive behaviors such as exploration, backtracking, and reasoning depth. These components increasingly recognized as crucial for developing reasoning LLMs~\citep{chen2025towards,gandhi2025cognitive,ye2025limo}. 
To our knowledge, our work is the first to explicitly extract this structural information and conduct a quantitative analysis of its correlation with reasoning quality.

The LCoT2Tree tool involves five automated stages that transform a LCoT %long chain of thought 
into an organized tree structure using an LLM (DeepSeek-v3;~\citealp{liu2024deepseek}), as shown in Figure~\ref{fig:sec3.2:lcot2tree}:

\noindent\textbf{Stage 1: Extract Sketch.} Leveraging the LLM with prompting (Figure~\ref{appendix:prompt1}), we condense the LCoT into a concise \textit{Reasoning Sketch} that outlines its main reasoning steps. The sketch serves as an abstract summary, highlighting the essential components and logical flow of the reasoning process.

\noindent\textbf{Stage 2: Split Thought.} 
%In this stage,  
%The LCoT is segmented into a list of thoughts. 
We first define a ``\textit{Thought}'' as a consecutive segment in the reasoning chain that involves no logical transition (\textit{e.g.}, exploration and verification).
%Then we analyze the collected LCoTs and identify common separators that signal transitions between distinct thoughts. 
%To carry out this segmentation, we identify common linguistic cues (\textit{i.e.}, separators) that signal a shift between reasoning steps. These separators include phrases such as ``Wait'', ``Alternatively'', and ``Let me verify''. Using these separators, we divide the the full chain into distinct reasoning fragments, producing a \textit{Thoughts List}, where each entry corresponds to one coherent unit of thought.
We utilize common linguistic cues (\textit{e.g.}, ``Wait'', ``Alternatively'', and ``Let me verify'') indicative of transitions between reasoning steps to segment the full reasoning chain into distinct fragments, yielding a \textit{Thoughts List}. 

\noindent\textbf{Stage 3: Assign Step.} Each thought in the \textit{Thoughts List} is matched to one or more steps in the \textit{Reasoning Sketch}, depending on its role in the overall reasoning process. This alignment is carried out using an LLM with prompting (Figure~\ref{appendix:prompt2}), generating a \textit{Thought Step} dictionary that maps each thought to its corresponding reasoning depths.

\noindent\textbf{Stage 4: Identify Function.} 
By prompting the LLM (Figure~\ref{appendix:prompt3}), we analyze consecutive thought pairs to determine the later thought's role relative to the former, with possible roles: (1) Continuous Logic; (2) Exploration; (3) Backtracking; and (4) Verification.
This assigns a \textit{Thoughts Function} label to each thought for clearer reasoning-flow purpose understanding.
%We analyze each pair of consecutive thoughts using LLM with prompt~\ref{appendix:prompt3} to determine the role of the latter thought in relation to the former. The possible roles include: (1) Continuous Logic; (2) Exploration; (3) Backtracking; and (4) Verification. 
%This process results in a \textit{Thoughts Function} label for each thought, providing a more precise understanding of its purpose within the reasoning flow. 
%For example, 
%whether the thought is introducing a new line of reasoning or verifying a previous assumption.
%whether the thought serves to introduce a new line of reasoning or to validate a prior assumption.

\begin{table*}[]
\begin{center}
\begin{small}
\begin{tabular}{l|c|ccccc|c}
\toprule
 &  & MATH & GPQA & LiveCodeBench & MMLU-Pro & 4 Datasets & Average \\ \midrule
\multirow{3}{*}{DeepSeek-32B} & Length-based & 74.13\% & 67.08\% & 81.59\% & 59.95\% & 66.27\% & 69.80\% \\
 & Tree-based & 80.81\% & 70.37\% & 82.21\% & 72.41\% & 71.14\% & 75.39\% \\
 & Gain & \textbf{+6.68\%} & \textbf{+3.29\%} & \textbf{+0.62\%} & \textbf{+12.46\%} & \textbf{+4.87\%} & \textbf{+5.58\%} \\ \midrule
\multirow{3}{*}{QwQ-32B} & Length-based & 75.82\% & 62.09\% & 78.30\% & 58.00\% & 66.97\% & 68.24\% \\
 & Tree-based & 77.63\% & 68.55\% & 80.05\% & 72.58\% & 70.98\% & 73.96\% \\
 & Gain & \textbf{+1.81\%} & \textbf{+6.46\%} & \textbf{+1.75\%} & \textbf{+14.58\%} & \textbf{+4.01\%} & \textbf{+5.72\%} \\ \midrule
\multirow{3}{*}{DeepSeek-R1} & Length-based & 76.94\% & 69.57\% & 81.75\% & 63.00\% & 70.54\% & 72.36\% \\
 & Tree-based & 80.30\% & 73.56\% & 81.80\% & 75.85\% & 73.72\% & 77.05\% \\
 & Gain & \textbf{+3.36\%} & \textbf{+3.99\%} & \textbf{+0.05\%} & \textbf{+12.85\%} & \textbf{+3.18\%} & \textbf{+4.69\%} \\ \midrule
\multirow{3}{*}{\begin{tabular}[c]{@{}l@{}}Seed-1.5-\\ Thinking-pro\end{tabular}} & Length-based & 67.48\% & 64.84\% & 76.39\% & 63.59\% & 64.34\% & 67.33\% \\
 & Tree-based & 70.07\% & 69.81\% & 77.72\% & 70.82\% & 67.68\% & 71.22\% \\
 & Gain & \textbf{+2.59\%} & \textbf{+4.97\%} & \textbf{+1.33\%} & \textbf{+7.23\%} & \textbf{+3.34\%} & \textbf{+3.89\%} \\\midrule
\multirow{3}{*}{\begin{tabular}[c]{@{}l@{}}Grok-3-\\ mini-beta\end{tabular}} & Length-based & 71.31\% & 61.48\% & 84.77\% & 55.47\% & 63.87\% & 67.38\% \\
 & Tree-based &  83.18\% & 66.79\% & 86.35\% & 70.68\% & 71.26\% & 75.65\% \\
 & Gain &  \textbf{+11.87\%} & \textbf{+5.31\%} & \textbf{+1.58\%} & \textbf{+15.21\%} & \textbf{+7.40\%} & \textbf{+8.27\%} \\ \bottomrule
\end{tabular}
\caption{Comparison of performance across various reasoning LLMs and datasets using the length-based method and our proposed tree-based approach for classifying response correctness based on LCoT information. Classification results are reported as the average over five runs. }
\label{tab:sec3.3:main_compare}
\vspace{-1em}
\end{small}
\end{center}
\end{table*}

\noindent\textbf{Stage 5: Build Tree.} 
%In this final stage, the segmented thoughts are organized into a hierarchical tree structure. 
Finally, we organize the segmented thoughts into a hierarchical tree structure.
Each node $N_i^j$ in the tree corresponds to the $i$-th thought $T_i$, where $j$ indicates how many times $T_i$ has appeared. The placement of a node is determined by the \textit{Thought Step}, and each edge represents a transition to a deeper level of reasoning, with the edge type defined by the \textit{Thought Function} of its child node. 
When inserting a new thought $T_i$, we first identify the ordered list of reasoning steps it maps to, denoted as $[S_i^1,...,S_i^n]$. 
% \yahui{I do not understand $S_i$ and how to compare it with $N_{i-1}$}. 
Here, $n$ indicates that the current thought encompasses $n$ reasoning steps. Consequently, we create $n$ nodes $N_i^1,...,N_i^n$, where each node $N_i^j$ represents the portion of the thought aligned with the $S_i^j$-th step. The insertion process follows two rules: (1)
%\begin{itemize}[leftmargin=*, nolistsep, noitemsep]
%\item 
If $S_i^1$ is greater than the step of the latest node $N_{i-1}^j$ in the tree, the new node $N_{i}^1$ is added as a child of $N_{i-1}^j$.
%\item 
(2) Otherwise, we backtrack to the most recent node at step $S_i^1-1$. Then we create a new branch from that node and link it to new node $N_{i}^1$. 
%\end{itemize}

Once $N_i^1$ is determined, the remaining nodes $N_i^2,...,$$N_i^n$ are added sequentially and connected to the previous one.
%, forming a linear path within the tree.
For example, in Figure~\ref{fig:sec3.2:lcot2tree}, when inserting $T_8$ to the tree, its associated reasoning steps $[S_8^1,S_8^2,S_8^3] = [1,2,3]$, as determined by the \textit{Thought Step}. At that point, the latest node in tree is $N_7^1$, which is at step 3 (greater than 1). Therefore, we backtrack to the latest node at step 0, $N_0^1$, and attach $N_8^1$ as its child. After that, $N_8^2$ and $N_8^3$ are linked sequentially to $N_8^1$ and $N_8^2$, respectively.

%In the end, we extract the whole tree structure using LCoT2Tree, which visually reveals how thoughts are connected and branch throughout the reasoning process.
In the end, we extract the tree structure, showing how thoughts are connected and branch throughout the reasoning process.
%This visual and structural representation allows researchers to understand and evaluate reasoning quality more accurately by considering not only the progression of steps, but also the complex pattern and depth of the underlying reasoning. 
% These tree structures highlight key cognitive patterns—such as exploration, reflection, and verification—and offer a more structured perspective on reasoning dynamics.
% We provide the detail prompt used in LCoT2Tree in the appendix~\ref{appendix:sec1}, along with several examples of the resulting visualized trees. 
% Researchers can assess not only the progression of steps but also the complexity and depth of the underlying thought patterns.
This structural representation offers three key benefits: (1) highlights key cognitive patterns (\textit{e.g.}, exploration, backtracking and verification); (2) supports more accurate assessment of reasoning quality; and (3) enables structure-aware analysis of reasoning behaviors. Implementation details, including prompts and case visualizations, are available in Appendix~\ref{appendix:sec1}.

%-----------------------------------%
\subsection{Effectiveness of LCoT2Tree}
\label{sec3.3}
%-----------------------------------%

To assess the effectiveness of the LCoT2Tree tool, we conduct a quantitative evaluation by using graph neural networks (GNNs) to predict answer correctness based on the tree structures extracted from LCoTs.
This evaluation demonstrates the practical value of tree-based representations for understanding complex reasoning processes.

\noindent\textbf{Experimental Setup.} We use the same dataset described in Section~\ref{sec3.1}, which contains responses from five reasoning models across four public benchmarks. 
The key difference is that we extract the tree structure from each LCoT response and use it as input to the GNNs. Our objective is to assess how effectively these tree structures can distinguish between correct and flawed reasoning. To this end, we utilize GATv2 \citep{brody2022how}, a GNN architecture suited for modeling hierarchical structures and their relationships. 
The model takes the nodes, edges, and associated features of each LCoT tree as input and learns a structural embedding that represents the overall reasoning pattern. Implementation details and graph construction are provided in Appendix~\ref{appendix:sec2}. We use classification accuracy as the evaluation metric. A high accuracy score indicates that the model successfully captures the correlation between reasoning structure and answer correctness. %, demonstrating the effectiveness of our LCoT2Tree framework in analyzing reasoning quality.

\noindent\textbf{Effectiveness across Tasks.}
Table~\ref{tab:sec3.3:main_compare} shows the classification results using tree-based input, compared to baseline methods that rely on the length-based feature. We assess how well the tree-based method generalizes across diverse types of reasoning tasks, including MATH, GPQA, LiveCodeBench, MMLU-Pro, and a combined dataset of these benchmarks. Across all tasks, the tree-based method consistently outperforms the length-based baseline. The improvement is particularly notable on MMLU-Pro, a dataset where reasoning correctness is difficult to predict from token length alone. For example, our method achieves substantial accuracy gains of +12.46\% and +14.58\% on DeepSeek-32B and QwQ-32B, respectively. Even on datasets like LiveCodeBench, where the length-based approach already performs strongly, the tree-based method still yields improvements, demonstrating its robustness.

\noindent\textbf{Effectiveness across Models.}
%To further assess the generalizability of our method, we evaluate five different models: DeepSeek-32B, QwQ-32B, Deepseek-R1, Seed-1.5-Thinking-pro, and Grok-3-mini-beta. 
For the generalizability of our method, 
the tree-based classifier consistently achieves higher accuracy than the length-based baseline across all models. Average accuracy gains range from +3.89\% (Seed-1.5-Thinking-pro) to +8.27\% (Grok-3-mini-beta), indicating that the LCoT2Tree provides more informative and reliable structural representations of reasoning processes.

%These findings suggest that answer correctness predictions of all models benefit from the structured representation introduced by LCoT2Tree, emphasizing the key role of reasoning structure in measuring reasoning quality.
% This findings demonstrate that the prediction of answer correctness for all models is enhanced by the structure representation provided by LCoT2Tree, underscoring the crucial role of the reasoning structure in evaluating reasoning quality. 

% These findings highlight the effectiveness of our method in capturing deeper cognitive patterns and improving reasoning quality evaluation across diverse tasks and model families.

These quantitative evaluation validates the effectiveness of the LCoT2Tree tool across diverse tasks and models. 
By capturing deeper structural and cognitive patterns in reasoning, it enables more accurate prediction of reasoning success. 
% Results highlight the effectiveness of our method in capturing deeper cognitive patterns and improving reasoning quality evaluation
% Results indicate that the tree structure derived from reasoning processes offer a more robust and detailed analysis of reasoning quality than traditional length-based methods. 
Overall, LCoT2Tree shows strong potential as an automated tool for analyzing, evaluating, and improving the behavior of reasoning systems.
%----------------------%
\section{Understand Behaviors of Reasoning Large Language Models}
\label{sec4}
%----------------------%

In this section, we leverage LCoT2Tree to analyze and understand reasoning behaviors. 
First, we identify key thought patterns in the reasoning tree that predict errors.
%We start by identifying key thought patterns within the reasoning tree that contribute to error prediction outcomes. 
%We then extend our analysis to distinguish reasoning behaviors across different tasks and models.
Then, we compare behaviors across tasks and models.
%Our findings reveal how reasoning behaviors differ by: (1) output correctness, (2) task type, and (3) model variant, highlighting the importance of structural information in reasoning analysis.
Our findings show that reasoning varies by (1) output correctness, (2) task type, and (3) model variant, underscoring the importance of structural information in reasoning analysis.

\noindent \textbf{Explainability Method.} To interpret the model’s predictions on reasoning quality and uncover the influential reasoning patterns, we adapt a graph explainability method called \textit{GNNExplainer}~\citep{ying2019gnnexplainer}.
This method uncovers important subgraphs by maximizing the mutual information between the GNN’s output and the distribution of possible subgraph structures. These extracted subgraphs also correspond to critical thought patterns within the reasoning chain.
For example, in models trained to predict incorrect answers, the highlighted subgraphs often reflect flawed reasoning behaviors that lead to poor performance. Similarly, in models trained on MATH tasks, the important subgraphs typically capture common reasoning patterns observed in mathematical problem-solving.

\begin{table*}[t]
\begin{center}
\begin{small}
\begin{tabular}{l|cccc | cc}
\toprule
\multirow{2}{*}{} & \multicolumn{4}{c}{\textbf{Task-specific Analysis}} & \multicolumn{2}{|c}{\textbf{Model-specific Analysis}} \\ \cmidrule(lr){2-5} \cmidrule(lr){6-7}
 &  MATH/GPQA & MATH/LCB & MATH/MMLU & GPQA/LCB & DS-32/DS-R1 & DS-32/Grok \\ \midrule
 Length-based  &  50.45\% & 63.72\% & 69.43\% & 60.65\% & 55.17\% & 61.06\% \\
 Tree-based & 83.51\% & 89.22\% & 78.46\% & 85.55\%   & 67.88\% & 93.22\% \\
 Gain & \textbf{+33.06\%} & \textbf{+25.50\%} & \textbf{+9.03\%} & \textbf{+24.90\%} & \textbf{+12.71\%} & \textbf{+32.16\%} \\  \bottomrule
\end{tabular}
\caption{ Comparison of task-specific and model-specific classification accuracy using the length-based method and the proposed tree-based approach. Task-specific analysis is conducted on the DeepSeek-32B model across different datasets, while model-specific analysis is performed on the MATH dataset across multiple model variants.}
\vspace{-1em}
\label{tab:sec4.2:task_split}
\end{small}
\end{center}
\end{table*}

\begin{figure}[t]
  \includegraphics[width=\columnwidth]{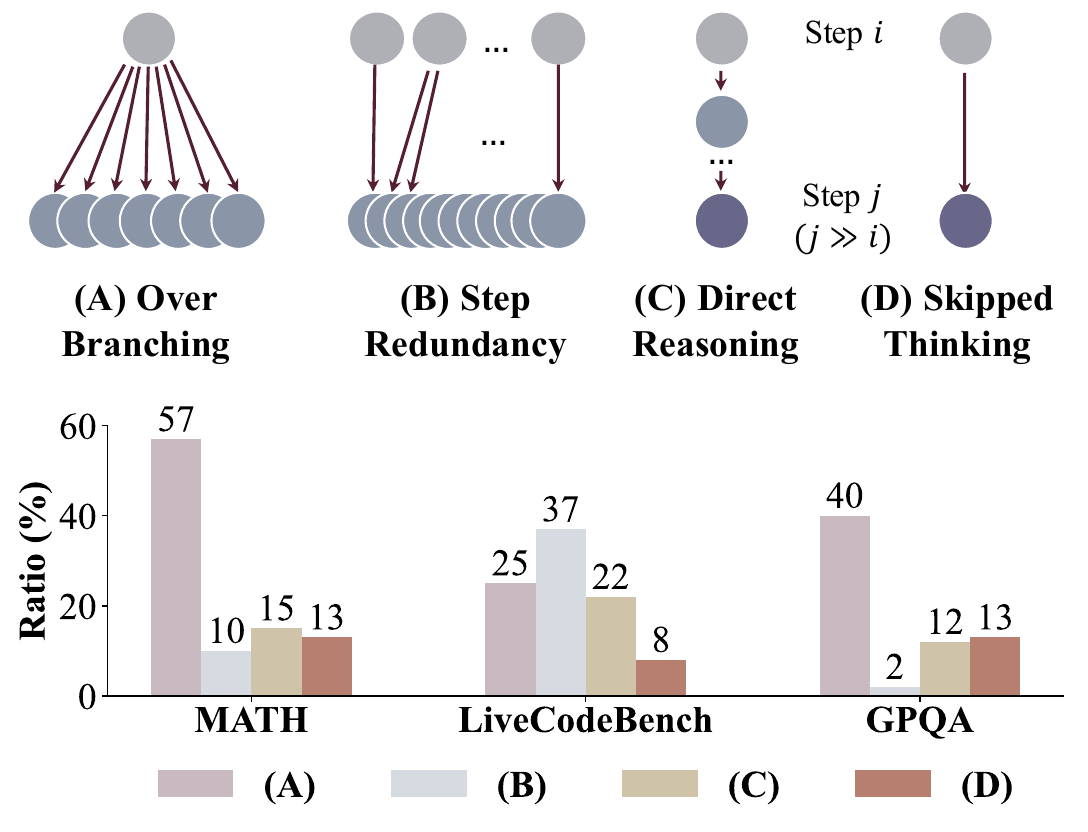}
  \vspace{-1em}
  \caption{Visualization and frequency of four structural error patterns across three datasets. (A) \textit{Over Branching}: abundance of explorations or verifications within a single node, (B) \textit{Step Redundancy}: over-generation of thoughts within a single reasoning step, (C) \textit{Direct Reasoning}: following a straight, minimal-branch path from one step to a much deeper step, and (D) \textit{Skipped Thinking}: jumping multiple steps ahead without intermediate logical analysis. Representative example of each class is available in Figure~\ref{appendix:visual_error_case}.}
  \vspace{-1em}
  \label{fig:sec4.1:error_pattern}
\end{figure}

\subsection{Error Patterns in LCoT}
\label{subsec:4.1_error_patterns}
The experiments in Section~\ref{sec3.3} suggest that reasoning trees of model responses exhibit separable structures for correct and incorrect outcomes. 
To further explore the behaviors that contribute to failures, we employ GNNExplainer to identify the most influential edges in each reasoning tree.  This allows us to extract critical subgraphs from incorrect responses and summarize common patterns across diverse examples (See details in Appendix~\ref{appendix:sec4.0}). The identified error patterns are visualized in the top portion of Figure~\ref{fig:sec4.1:error_pattern}, with detailed examples shown in Figure~\ref{appendix:visual_error_case}. Additionally, we analyze 100 error responses from three tasks and report the frequency of each pattern in the bottom portion of Figure~\ref{fig:sec4.1:error_pattern}. A key observation is that excessive and insufficient branching are both strongly associated with incorrect reasoning.
% It indicates that over-branching at a specific reasoning step not only complicates the reasoning process but also increases the probability of errors. 

%-------------------------------%
\subsection{Task-Specific Patterns in LCoT}
\label{subsec:4.2_task_specific}
%-------------------------------%
In the left part of Table~\ref{tab:sec4.2:task_split}, we present the results of a task separability experiment conducted on the DeepSeek-32B model. We classify reasoning trees across task pairs (\textit{e.g.}, MATH/GPQA, MATH/LCB, MATH/MMLU-Pro, GPQA/LCB), with additional results for QwQ-32B provided in Appendix~\ref{appendix:sec3.2}. The dataset follows the same construction as in Section~\ref{sec3.1}, but labels tasks instead of correctness.
Results show that our tree-based method effectively distinguishes task-specific reasoning patterns, achieving an average accuracy of 84.19\%. Notably, in cases where length-based features fall short, such as MATH/GPQA and GPQA/LCB, tree-based representations yield substantial gains of +33.06\% and +24.90\%, respectively, highlighting their strength in capturing deeper reasoning patterns.

\noindent\textbf{Discovering Task-Specific Reasoning Patterns.} Beyond quantitative separation, we further leverage LCoT2Tree to reveal task-specific reasoning patterns through qualitative analysis. Main conclusions, based on DeepSeek-32B, are as follows: For MATH (Figure~\ref{appendix:visual_ds32_math}), the reasoning trees exhibit a diagonally descending structure, reaching deeper steps through repeated backtracking. This reflects a layered, step-by-step problem-solving approach. In contrast, the behaviors in code completion (Figure~\ref{appendix:visual_ds32_code}) show wide, parallel branches with minimal exploration or verification, reflecting a more straightforward pattern of generation. GPQA (Figure~\ref{appendix:visual_ds32_gpqa}) samples reveal high out-degree nodes, where the model repeatedly revisits complex concepts,  indicating the model’s uncertainty in dealing an expert-level question. Meanwhile, the trees of MMLU-Pro (Figure~\ref{appendix:visual_ds32_mmlu}) are relatively shallow with minimal branching, reflecting a straightforward deductive reasoning style that aligns with the nature of knowledge-based questions. These observations highlight LCoT2Tree’s ability to provide interpretable insights into the distinct reasoning strategies employed across different task types. Detailed case studies are provided in Appendix~\ref{appendix:sec4.1} with visualizations shown in Figure~\ref{appendix:visual_ds32_math} - Figure~\ref{appendix:visual_ds32_mmlu}.

\subsection{Model-Specific Patterns in LCoT}
\label{subsec:4.3_model_specific}
We explore whether different models exhibit distinguishable reasoning behaviors on the same dataset. The results, shown in the right part of Table~\ref{tab:sec4.2:task_split}, demonstrate that LCoT2Tree effectively captures model-specific patterns. In particular, tree-based representations significantly outperform simple length-based features, with gains of +12.71\% for DS-32 (DeepSeek-32B) vs. DS-R1 (DeepSeek-R1), and +32.16\% for DS-32 vs. Grok (Grok-3-mini-beta). 
Notably, the relatively lower separability score between DS-32 and DS-R1 (67.88\%) can be attributed to the fact that DS-32 is a distilled version of DS-R1. In contrast, DS-32 and Grok show a high separability of 93.22\%, suggesting fundamentally different reasoning styles driven by architectural and training differences. Additional results (Appendix~\ref{appendix:sec3.3}) show that QwQ-32B aligns more closely with the DeepSeek family than with Grok or Seed (Seed-1.5-Thinking-pro). 
These findings again highlight the strength of structural representations in revealing fine-grained behavioral distinctions.
% For example, QwQ-32B and DeepSeek-32B have extremely similar reasoning patterns, that is, behaviors such as exploration, verification, and reflection during the reasoning process will consistently occur.
% We speculate that it may be due to the relatively small differences in the training strategies, data, and other aspects of the models.
%This overlap complicates classification as their reasoning structures are indistinguishable.
% This significant overlap in reasoning structures poses a challenge to classification, rendering it difficult to differentiate between these models.

\noindent\textbf{Discovering Model-Specific Reasoning Patterns.} To complement the quantitative analysis, we also conduct a qualitative comparison of reasoning trees across different models on the MATH dataset (Appendix~\ref{appendix:sec4.2}). Our analysis reveals that both DS-R1 (Figure~\ref{appendix:visual_dsr1_math}) and QwQ-32B (Figure~\ref{appendix:visual_qwq_math}) produce reasoning structures similar to DS-32 (Figure~\ref{appendix:visual_ds32_math}), consistent with quantitative results. However, DS-R1 tends to prune its reasoning paths earlier, suggesting a more aggressive backtracking strategy. In contrast, QwQ-32B shows more extensive exploration in the later stages of reasoning. On the other hand, Seed (Figure~\ref{appendix:visual_seed_math}) and Grok (Figure~\ref{appendix:visual_grok_math}) follow simpler, more linear reasoning paths with fewer thought transitions and minimal branching, reflecting a straightforward reasoning strategy.

\subsection{Shortcomings in Understanding LCoT from the Structural Perspective}
\label{sec4.4}

\noindent\textbf{Correct Structure but Wrong Output.} %\yahui{add a example in the appendix}
Despite structurally valid reasoning paths, models can still produce incorrect answers due to semantic errors like misinterpreting the problem, making calculation mistakes, or failing conditional logic. This indicates that reasoning LLMs do not consistently exhibit behaviors like backtracking or verification when facing ambiguity or errors. These cases expose the limitation of structural analysis alone and suggests that combining structural insights with semantic verification is necessary for comprehensive reasoning understanding.

% A key assumption in predicting answer correctness from structure is that reasoning LLMs will reliably exhibit behaviors such as backtracking or verification when faced with semantic errors or ambiguities. 
% However, we find cases where the structure appears valid, but reasoning contains semantic flaws—such as misinterpreting the problem, making calculation mistakes, or failing conditional logic—that ultimately lead to incorrect answers. 
% This reveals the limits of structural perspectives and suggests that optimizing reasoning through structure reflects internal model improvements, whereas incorporating semantic verifiers offer complementary, external optimization.
% and the need to incorporate semantic validation in evaluating reasoning quality.
% We find instances where tree structures are predicted as correct, but semantic errors in reasoning step lead to incorrect final answers. 
% For instance, the reasoning trees of specific math problems incorporate correct computational steps; however, errors emerge due to the misapplication of mathematical principles.
% This suggests that although the structure can reflect quality to a certain degree, it is insufficient to comprehensively assess semantic accuracy. 
% However, we can still observe that Long CoT with a correctly structured pattern is more likely to at least ``seem'' to be correct.

\noindent\textbf{Flawed Structure but Correct Output.} 
%\yahui{add a example in the appendix}
%Conversely, some answers are final answer correct but involved unconventional or flawed reasoning structures. 
%We identify a subset of responses where the final answer is correct despite weak or flawed reasoning, as detected by our classifier. 
Using our classifier, we identify a set of responses with correct final answer but weak or flawed reasoning. 
These cases often involve reasoning paths that deviate from systematic problem-solving, including guessing, brute-force enumeration, or overly late-stage corrections. 
Such cases underscore the limitations of using answer correctness alone to assess reasoning quality, as it tolerates shallow or unsound reasoning paths. 
Addressing these flawed-but-correct patterns can guide LLMs toward producing reasoning that is not only accurate but also logically sound.

% Overall, while LCoT2Tree provides unique insights into reasoning patterns, allowing for a detailed analysis of hierarchical reasoning structures, error sources, and task- or model-specific behaviors, it still has clear limitations. It fails to detect semantic errors or unconventional shortcuts that deviate from the established structural expectations.
% For future endeavors, 
% %Future work will combine semantic reasoning evaluation with structural analysis for a more holistic understanding of reasoning behaviors in LLMs.
% the integration of semantic reasoning evaluation and structural analysis will be carried out to obtain a more all-encompassing view of the reasoning behaviors in LLMs.

\section{Application of LCoT2Tree: Tree-based Best-of-N Decoding}
\label{sec5}
%Leveraging the insights provided by LCoT2Tree, we propose practical applications that can aid both the decoding and evaluation of LLMs. 
% Building on the valuable insights offered by LCoT2Tree, we put forward practical applications that can effectively assist in the decoding process of LLMs.
% Specifically, %we demonstrate how LCoT2Tree's structural understanding can enhance decoding strategies through tree-based Best-of-N decoding and establish a robust reasoning benchmark for LLM evaluation. 
% we illustrate how the structural comprehension provided by LCoT2Tree can improve decoding strategies by means of tree-based Best-of-N decoding. 
% and  serve as a foundation for constructing a robust reasoning benchmark to evaluate LLM. 

% \subsection{Tree-based Best-of-N Decoding}

%Building on the demonstrated effectiveness of LCoT2Tree in 
Beyond evaluating the quality of model reasoning (Section~\ref{sec3.3}), we put forward practical application to support the decoding process in LLMs. Specifically, we propose an approach to improve the reasoning quality during the decoding stage by selecting the best model response from multiple candidates with the tree-based classifier.

\noindent\textbf{Method.}
Best-of-N decoding is a widely used strategy for improving the quality of responses generated by LLMs~\citep{wu2024scaling,snell2024scaling,brown2024large}. 
In this strategy, the model produces $N$ candidate outputs, and a final response is selected based on a scoring function. However, conventional scoring methods, based on surface-level heuristics or reward models, often ignore the impact of output structures. This limitation can lead to suboptimal choices, especially in tasks that require deep or structured reasoning.

To this end, we incorporate LCoT2Tree into the Best-of-N decoding framework to guide the selection of high-quality reasoning outputs. Our method involves three main steps:
(1) For each candidate response, we use LCoT2Tree to build its corresponding reasoning tree; (2) A graph-based classifier, trained to distinguish between successful and flawed reasoning structures, assigns a score to each candidate based on its structural features; (3) The candidate with the highest score is chosen as the final output.
% By emphasizing the structural integrity of reasoning, this tree-based selection process leads to more accurate and coherent results, especially in tasks that require complex, multi-step reasoning.\yahui{delete the last sentence?}

\noindent\textbf{Experiments.} We choose LiveCodeBench (LCB) as our primary benchmark. We train the GNN models following the setup in Section~\ref{sec3.3} using LCB-v5 dataset and then evaluate on a challenging subset (filtered by correctness ratio) of the LCB-v6 dataset. We compare our tree-based Best-of-N decoding method with three baselines: (1) \textit{ORM-Best}~\citep{brown2024large}, which selects the response with the highest score from an outcome reward model (we use Skywork-Reward-Gemma-2-27B-v0.2~\citep{liu2024skywork}); (2) \textit{PRM-Best}, which scores responses based on the product of step-level scores from a process reward model (\textit{i.e.}, Qwen2.5-Math-PRM-72B~\citep{zhang2025lessons}); %following the procedure in; 
and (3) \textit{Length-Best}~\citep{wang2025thoughts}, which selects the response with the fewest tokens. All experiments use $N = 10$ candidate responses. Additional results on MATH with more baselines are presented in Appendix~\ref{appendix:sec3.4}.

\begin{figure}[t]
  \includegraphics[width=\columnwidth]{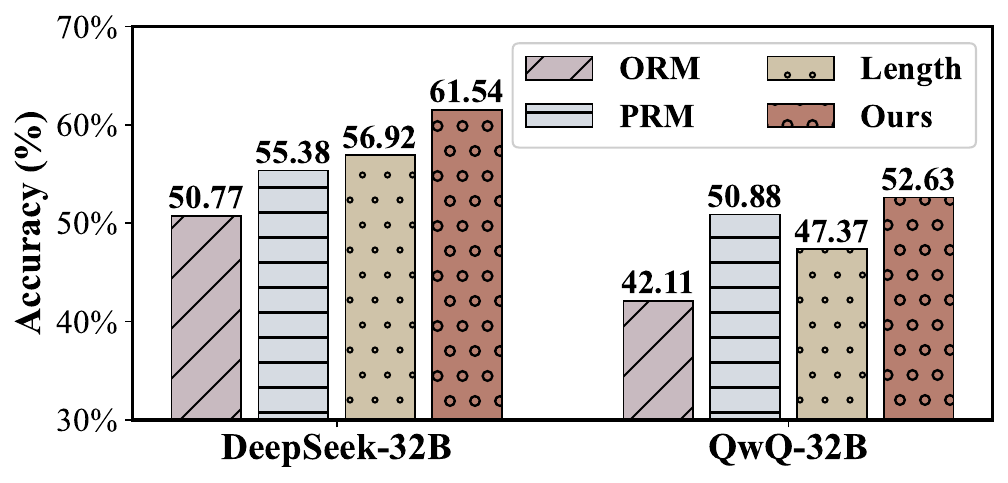}
  \vspace{-1em}
  \caption{Accuracy comparison of different Best-of-N decoding strategies on the LCB-v6 benchmark.}
  \vspace{-1em}
  \label{fig:sec5.1:best-of-n}
\end{figure}

% \begin{table}[]
% \begin{center}
% \begin{small}
% \begin{tabular}{@{}l|l|cc|c@{}}
% \toprule
%  &  & Length-Best & ORM-Best &  Ours-Best\\ \midrule
% \multirow{2}{*}{LCB}  &  DS-32 &  56.92\%  & 50.77\%  & \textbf{61.54\%} \\
%  & QwQ-32  & 47.37\% & 42.11\%  & \textbf{52.63\%} \\ \bottomrule
% \end{tabular}
% \caption{Accuracy comparison of different Best-of-N decoding strategies on the LCB-v6 benchmark.}
% \label{tab:sec5.1:best_of_n}
% \end{small}
% \end{center}
% \end{table}

\noindent\textbf{Results.} As shown in Figure~\ref{fig:sec5.1:best-of-n}, our tree-based Best-of-N method outperforms both Length-Best, ORM-Best and PRM-Best on the LCB-v6 benchmark. For DeepSeek-32B, it achieves 61.54\% accuracy, exceeding Length-Best by +4.62\%, ORM-Best by +10.77\% and PRM-Best by +6.16\%. QwQ-32B shows similar gains, with our method reaching 52.63\%, outperforming the baselines by +5.26\%, +10.52\% and 1.75\%, respectively. These results highlight the advantage of using structural reasoning signals via LCoT2Tree to improve candidate selection in complex tasks.

% \subsection{Reasoning LLM Robust Evaluation Benchmark}
% \noindent\textbf{Method}

% \noindent\textbf{Benchmark Statistics}

% \noindent\textbf{Results}

% \begin{table}[]
% \begin{center}
% \begin{small}
% \begin{tabular}{@{}l|cc|cc@{}}
% \toprule
%  & MATH500 & *TPert & GPQA & *TPert \\ \midrule
% Deepseek-32b &  &  &  &  \\
% QwQ-32b &  &  &  &  \\
% Deepseek-R1 &  &  &  &  \\
% Doubao-1.5 &  &  &  &  \\
% Grok-3-mini &  &  &  &  \\ \bottomrule
% \end{tabular}
% \caption{}
% \label{tab:sec5.2:robustness}
% \end{small}
% \end{center}
% \end{table}
\section{Conclusion}

In this work, we introduce a novel framework, named LCoT2Tree, for converting LCoT responses into hierarchical tree structures. 
%This representation 
% The proposed 
% LCoT2Tree enables more interpretable and structural analysis of complex reasoning processes. 
% Through quantitative experiments, we demonstrate that the tree-based representations significantly improve the prediction of reasoning quality across a wide range of tasks and models. 
LCoT2Tree enables more interpretable and structural analysis of complex reasoning processes, with significantly improving the prediction of reasoning success across a wide range of tasks and models. 
Beyond evaluation, we apply LCoT2Tree for behavioral analysis, revealing error patterns and accounting for disparate behaviors across tasks and models. 
Furthermore, we extend LCoT2Tree to a practical application by integrating it into the Best-of-N decoding paradigm, leading to more accurate outputs than ORM, PRM and length-based baselines. 
Collectively, these findings underscore the significance of structural reasoning analysis and establish LCoT2Tree as a promising tool for understanding and improving LLMs reasoning capabilities.

\section*{Limitations}
While LCoT2Tree is a powerful framework for analyzing reasoning structures, several limitations remain. First, as discussed in Section~\ref{sec4.4}, structural analysis alone cannot capture semantic errors or recognize correct reasoning that deviates from common structural patterns. To address this limitation, future work should integrate semantic reasoning signals with structural analysis to achieve a more holistic understanding of LLM reasoning behaviors.

% Second, one key reason we are able to assess reasoning success via structural patterns is that current LLMs often exhibit incomplete or loosely structured reasoning. 
% %that is incomplete or loosely structured. 
% As LLMs continue to evolve and produce more coherent and well-organized reasoning, the value of structural analysis may diminish. However, until that point, leveraging structural insights remains a promising approach for improving model reasoning.

Second, the effectiveness of structural analysis relies in part on the fact that current LLMs often generate reasoning that is incomplete or loosely organized. As models improve and begin to produce more coherent and well-structured reasoning by default, the value of structural diagnostics may decrease. Nevertheless, until such consistency is achieved, structural cues remain a valuable tool for identifying and improving reasoning quality.

Finally, the construction of reasoning trees in LCoT2Tree currently depend on \textit{off-the-shelf} large language models (e.g., DeepSeek-V3~\citep{liu2024deepseek}), which makes the pipeline computationally expensive.
% Finally, LCoT2Tree's
% %enhances interpretability,  
% tree construction and analysis rely on the \textit{off-the-shelf} large language models (\textit{e.g.}, DeepSeek-V3~\citep{liu2024deepseek}), making the current process computationally expensive.

% \section*{Acknowledgments}

\bibliography{custom}

\appendix

\label{sec:appendix}

\clearpage

%---------------------------------%
\section{LCoT2Tree Tool Implementation Details}
\label{appendix:sec1}
%---------------------------------%
The LCoT2Tree process involves five automated stages to transform a LCoT into an organized tree structure using the LLM (DeepSeek-v3;~\citealp{liu2024deepseek}), as shown in Figure~\ref{fig:sec3.2:lcot2tree}. Here, we introduce the detailed implementation of each step:

\noindent\textbf{Stage 1: Extract Sketch.} Leveraging the LLM with prompt~\ref{appendix:prompt1}, we condense the LCoT into a sketch that captures its core reasoning steps. This \textit{Reasoning Sketch} provides an abstract of the reasoning process, focusing on the key steps and the logical flow of the reasoning. 

\noindent\textbf{Stage 2: Split Thought.} In this stage,  the LCoT is split into a list of thoughts. We first define a ``\textit{Thought}'' as: a continue segment in a reasoning chain that involves no logical transition, such as exploration or verification. 
%Then we analyze the collected LCoTs and identify common separators that signal transitions between distinct thoughts. 
We then analyze the collected LCoTs to identify common linguistic patterns (\textit{i.e.}, separators) that signal shifts between distinct reasoning steps. The separators set to [``Alternatively'', ``Hmm'', ``Let me verify'', ``let's verify'', ``To verify'', ``Wait'', ``Verify''] for Deepseek-32B, QwQ-32B, and Deepseek-R1. And extend with [``Let's confirm'', ``Let's check'', ``Another example'', ``But let's'', ``wait'', ``No:'', ``no:'', ``Now''] for Seed-1.5-thinking-pro and Grok-3-mini-beta. According to these markers, the long reasoning chain is divided into individual thoughts, forming a \textit{Thoughts List} where each item represents a single reasoning fragment.
    
\noindent\textbf{Stage 3: Assign Step.} Each thought in the \textit{Thoughts List} is then aligned with one or more corresponding steps in the \textit{Reasoning Sketch}, based on its role in the overall reasoning process. This mapping is performed using an LLM with prompt~\ref{appendix:prompt2}, generating a \textit{Thought Step} dictionary that captures the contextual meaning and reasoning stage (\textit{i.e.}, depth) associated with each thought. To improve token efficiency, we group and merge adjacent thoughts before feeding them into the LLM. However, due to the large number of thoughts in the \textit{Thought List}, processing them all at once is infeasible. Therefore, we segment the list into smaller batches, each containing consecutive thoughts with a combined word count of no more than 600. These batches are then input to the LLM, which returns the corresponding reasoning step for each thought in a single response.
    
\noindent\textbf{Stage 4: Identify Function.} We further analyze each pair of consecutive thoughts using LLM with prompt~\ref{appendix:prompt3} to determine the role of the latter thought in relation to the former (\textit{e.g.}, continuation, exploration, or verification). This step provides a more precise understanding of the relationships between individual thoughts within the reasoning process. Specifically, the roles are categorized as follows: (1) Continuous Logic – A direct continuation or extension of the reasoning in the previous thought. (2) Exploration – Introduces alternative reasoning paths, unrelated concepts, or new topics. (3) Backtracking – Revises, corrects, or adjusts the reasoning from the previous step. (4) Validation – Provides supporting evidence, justification, or examples for the previous thought. If the \textit{Thought List} contains $N$ thoughts, we perform $N-1$ LLM calls to analyze each adjacent pair.

\noindent\textbf{Stage 5: Build Tree.} 
%In this final stage, the segmented thoughts are organized into a hierarchical tree structure. 
Finally, we organize the segmented thoughts into a hierarchical tree structure.
Each node $N_i^j$ in the tree corresponds to the $i$-th thought $T_i$, where $j$ indicates how many times $T_i$ has appeared. The placement of a node is determined by the \textit{Thought Step}, and each edge represents a transition to a deeper level of reasoning, with the edge type defined by the \textit{Thought Function} of its child node. 
When inserting a new thought $T_i$, we first identify the ordered list of reasoning steps it maps to, denoted as $[S_i^1,...,S_i^n]$. 
% \yahui{I do not understand $S_i$ and how to compare it with $N_{i-1}$}. 
Here, $n$ indicates that the current thought encompasses $n$ reasoning steps. Consequently, we create $n$ nodes $N_i^1,...,N_i^n$, where each node $N_i^j$ represents the portion of the thought aligned with the $S_i^j$-th step. The insertion process follows two rules: (1)
%\begin{itemize}[leftmargin=*, nolistsep, noitemsep]
%\item 
If $S_i^1$ is greater than the step of the latest node $N_{i-1}^j$ in the tree, the new node $N_{i}^1$ is added as a child of $N_{i-1}^j$.
%\item 
(2) Otherwise, we backtrack to the most recent node at step $S_i^1-1$. Then we create a new branch from that node and link it to new node $N_{i}^1$. 
%\end{itemize}

Once $N_i^1$ is placed, the remaining nodes $N_i^2,...,$$N_i^n$ are added sequentially and connected to the previous one.
%, forming a linear path within the tree.
For example, in Figure~\ref{fig:sec3.2:lcot2tree}, when inserting $T_8$ to the tree, its associated reasoning steps $[S_8^1,S_8^2,S_8^3] = [1,2,3]$, as determined by the \textit{Thought Step}. At that point, the latest node in tree is $N_7^1$, which is at step 3—greater than 1. Therefore, we backtrack to the latest node at step 0, $N_0^1$, and attach $N_8^1$ as its child. After that, $N_8^2$ and $N_8^3$ are linked sequentially to $N_8^1$ and $N_8^2$, respectively.

In the end, we successfully extract the whole tree structure using LCoT2Tree. To support interpretation, we provide a visualization tool for the generated reasoning tree, which allows users to interactively explore the thought process behind each node while viewing the overall tree structure. Example screenshots of the visualization results are shown in Figures~\ref{appendix:visual_ds32_math}–\ref{appendix:visual_grok_math}. In each figure, the right column displays the key reasoning steps identified in Step 1, and each node represents an individual thought. The solid line denotes the edge from father to child and the dash line denotes the edge from child to father. Edges are colored according to its function in reasoning process.
% Figures~\ref{appendix:visual_pos_1} and~\ref{appendix:visual_pos_2} show trees generated from correctly answered samples, while the remaining figures illustrate cases where the responses were incorrect. 
%  We observe that incorrect samples often display excessive and repetitive exploration at certain nodes, in contrast to the more streamlined structure of correct ones. This pattern may partially explain the drop in performance associated with those responses.

\begin{table*}[]
\begin{center}
\begin{small}
\begin{tabular}{l|l|cccc cc}
\toprule
 & &  MATH/GPQA & MATH/LCB & MATH/MMLU & GPQA/LCB & GPQA/MMLU & MMLU/LCB \\ \midrule
\multirow{3}{*}{DS-32} & Length-based  &   50.45\% & 63.72\% & 69.43\% & 60.65\% & 77.71\% & 82.34\% \\
& Tree-based & 83.51\% & 89.22\% & 78.46\% & 85.55\% & 82.89\% & 92.12\% \\
& Gain & \textbf{+33.06\%} & \textbf{+25.50\%} & \textbf{+9.03\%} & \textbf{+24.90\%} & \textbf{+5.18\%} & \textbf{+9.78\%} \\ \midrule
\multirow{3}{*}{QwQ-32} & Length-based  &  52.51\% & 56.64\% & 67.38\% & 61.22\% & 68.25\% & 73.03\% \\
& Tree-based & 77.82\% & 67.88\% & 80.85\% & 85.69\% & 76.70\% & 87.20\% \\
& Gain &\textbf{+25.31\%} & \textbf{+11.24\%} & \textbf{+12.17\%} & \textbf{+24.47\%} & \textbf{+8.45\%} & \textbf{+14.17\%} \\  \bottomrule
\end{tabular}
\caption{Comparison of task-specific classification accuracy using the baseline length-based method and the proposed tree-based representation.}
\label{appendix:tab:task_split}
\end{small}
\end{center}
\end{table*}

\section{Classification Implementation Detail}

\label{appendix:sec2}

\subsection{Dataset Construction}
We use the same dataset as described in Section~\ref{sec3.1}, consisting of response samples generated by five reasoning LLMs (LLMs) across four public benchmarks: MATH, GPQA, LiveCodeBench (LCB), and MMLU-Pro. Each sample is labeled as positive or negative with answer correctness, which serves as the ground truth for our binary classification task. To ensure sufficient data volume, we apply repeated sampling for each benchmark, generating up to 2,000 samples per dataset. For instance, the LCB benchmark contains 167 unique problems. By generating 16 responses per problem, we obtain approximately 1,000 correctly answered samples and 1,000 incorrect ones.

\subsection{Graph Construction}
For each response, we begin by applying the LCoT2Tree framework to convert it into a structured reasoning tree. Each node $N_i^j$ in the tree corresponds to the $i$-th thought $T_i$, where $j$ indicates how many times $T_i$ has appeared. The placement of a node is determined by the \textit{Thought Step}, and each edge represents a transition to a deeper level of reasoning, with the edge type defined by the \textit{Thought Function} of its child node as introduced in~\ref{sec3.2}. We then transform the tree into a graph representation. Notably, we construct bidirectional edges, allowing information to flow both from parent to child and from child to parent. This design enables the model to simulate behaviors like backtracking, which are often essential in complex reasoning. In the end, each sample produces a single graph instance for classification.

\subsection{Node and Edge Features}
We design informative features for both nodes and edges to enhance the performance of our tree-based classification model. For each node in the reasoning tree, we extract the following features: (1) the index of the current thought, (2) the reasoning depth of the current node, (3) the cumulative number of tokens used up to current node, (4) the number of child nodes, and (5) the cumulative number of nodes at the same reasoning depth.

For edge features, we assign each parent-to-child edge a feature based on its logical role as identified by LCoT2Tree: ``1'' for continuation, ``2'' for exploration, ``3'' for backtracking, and ``4'' for validation. To distinguish child-to-parent edges (used to capture reverse information flow, such as backtracking), we assign the same value but multiply it by -1. This setup helps the model differentiate directional semantics during message passing.

\subsection{Hyperparameters}

We adopt the GATv2 architecture~\citep{brody2022how} to model reasoning trees, leveraging its dynamic attention mechanism and improved capability for capturing hierarchical dependencies. The model comprises two GATv2 layers, each with a hidden size of 64. After message passing, graph-level embeddings are obtained via global mean pooling. These embeddings are then fed into a two-layer MLP with ReLU activation, serving as the classification head to predict whether a given reasoning structure leads to a correct or incorrect answer.

To train the model, we use binary cross-entropy loss and the Adam optimizer with a learning rate of 1e-3. The model is trained for up to 100 epochs with a batch size of 32. We split the training dataset into 90\% for training and 10\% for validation. All experiments are conducted using the PyTorch Geometric framework.

\section{Additional Experimental Results}
\label{appendix:sec3}

\begin{table*}[]
\begin{center}
\begin{small}
\begin{tabular}{l|ccccc}
\toprule
 & MATH & GPQA & LiveCodeBench & MMLU-Pro & 4 Datasets  \\ \midrule
DeepSeek-32B & $\pm $0.0048& $\pm$ 0.0037 & $\pm$0.0024  & $\pm$ 0.0089& $\pm$ 0.0043\\ \midrule
QwQ-32B & $\pm$0.0023& $\pm$0.0025 & $\pm$0.0030 & $\pm$0.0079 & $\pm$ 0.0039\\ \midrule
DeepSeek-R1 & $\pm$0.0076& $\pm$0.0029 & $\pm$0.0010 &  $\pm$0.0051& $\pm$ 0.0018 \\\midrule
Seed-1.5-Thinking-pro  & $\pm$0.0037& $\pm$0.0061 & $\pm$0.0041& $\pm$0.0066& $\pm$ 0.0026 \\ \midrule
Grok-3-mini-beta& $\pm$0.0025& $\pm$0.0020 & $\pm$0.0037 & $\pm$0.0153& $\pm$ 0.0041\\ \bottomrule
\end{tabular}
\caption{Standard deviation of our proposed tree-based approach on classifying response correctness based on LCoT information corresponding to Table~\ref{tab:sec3.3:main_compare}.}
\label{appendix:tab:}
\vspace{-1em}
\end{small}
\end{center}
\end{table*}

\begin{table*}[]
\begin{center}
\begin{small}
\begin{tabular}{l|l|ccccc}
\toprule
 & &  DS-32/DS-R1 & DS-32/QwQ-32 & DS-32/Seed & DS-32/Grok & DS-R1/Seed \\ \midrule
\multirow{3}{*}{MATH} & Length-based  &    55.17\% & 61.49\% & 55.58\% & 61.06\% & 56.90\%\\
& Tree-based & 67.88\% & 70.93\% & 82.15\% & 93.22\% & 80.10\%\\
& Gain & \textbf{+12.71\%} & \textbf{+9.44\%} & \textbf{+26.57\%} & \textbf{+32.16\%} & \textbf{+23.20\%} \\ \midrule
\multirow{3}{*}{GPQA} & Length-based  & 50.87\% & 51.12\% & 67.96\% & 49.43\% & 65.34\%  \\
& Tree-based & 75.34\% & 61.60\% & 95.20\% & 99.42\% & 84.68\% \\
& Gain & \textbf{+24.47\%} & \textbf{+10.48\%} & \textbf{+27.24\%} & \textbf{+49.99\%} & \textbf{+19.34\%} \\ \midrule
\multirow{3}{*}{LCB} & Length-based  & 54.49\% & 54.17\% & 52.37\% & 54.49\% & 53.39\% \\
& Tree-based & 86.32\% & 71.73\% & 96.12\% & 86.32\% & 82.51\% \\
& Gain &\textbf{+31.83\%} & \textbf{+17.56\%} & \textbf{+43.75\%} & \textbf{+31.83\%} & \textbf{+29.12\%}\\ \midrule
\multirow{3}{*}{MMLU} & Length-based  &  55.36\% & 60.10\% & 54.17\% & 53.23\% & 59.55\%  \\
& Tree-based & 62.86\% & 64.99\% & 73.65\% & 85.62\% & 71.89\%  \\
& Gain & \textbf{+7.50\%} & \textbf{+4.89\%} & \textbf{+19.48\%} & \textbf{+32.39\%} & \textbf{+12.34\%} \\  \bottomrule
\end{tabular}
\caption{Comparison of model-specific classification accuracy using the baseline length-based method and the proposed tree-based representation.}
\label{appendix:tab:model_split}
\end{small}
\end{center}
\end{table*}

\begin{table*}[]
\begin{center}
\begin{small}
\begin{tabular}{l|cc|cc}
\toprule
 & \multicolumn{2}{c|}{DeepSeek-32B} & \multicolumn{2}{c}{QwQ-32B} \\ \midrule
 & LiveCodeBench & MATH & LiveCodeBench & MATH \\ \midrule
Vote & - & 80.41\% & - & \textbf{71.19\%} \\ \midrule
Length-Best & 56.92\% & 56.70\% & 47.37\% & 55.93\% \\ 
Length-Vote & - & 67.01\% & - & 57.63\% \\ \midrule
ORM-Best & 50.77\% & 60.82\% & 42.11\% & 57.63\% \\ 
ORM-Vote & - & 68.04\% & - & 67.80\% \\ \midrule
PRM-Best & 62.89\% & 63.92\% & 50.88\% & 57.63\% \\
PRM-Vote & - & 62.89\% & - & 55.93\% \\ \midrule
Ours-Best & \textbf{61.54\%} & 65.98\% & \textbf{52.63\%} & 67.80\% \\
Ours-Vote & - & \textbf{82.47\%} & - & \textbf{71.19\%} \\ \bottomrule
\end{tabular}
\caption{Accuracy comparison of different Best-of-N decoding strategies on the two benchmark.}
\label{appendix:tab:best_of_n}
\end{small}
\end{center}
\end{table*}

% \subsection{Additional Results on Token Length Distribution}
% \label{appendix:sec3.1}

\subsection{Additional Results on Task-specific Analysis}
\label{appendix:sec3.2}
In Table~\ref{appendix:tab:task_split}, we provide additional results from the task separability experiments using the DeepSeek-32B and QwQ-32B models. We classify reasoning trees across all task pairs, including MATH/GPQA, MATH/LCB, MATH/MMLU-Pro, GPQA/LCB, GPQA/MMLU-Pro, and LCB/MMLU-Pro. The findings are consistent with the conclusions presented in Section~\ref{subsec:4.2_task_specific}.

\subsection{Additional Results on Model-specific Analysis}
\label{appendix:sec3.3}
Table~\ref{appendix:tab:model_split} presents the detailed analysis of whether different models display distinguishable reasoning behaviors when applied to the same dataset. The results confirm that LCoT2Tree effectively captures model-specific reasoning patterns that generalize across tasks. Specifically, QwQ-32B exhibits reasoning behaviors more closely aligned with the DeepSeek family, compared to Grok-3-mini-beta and Seed-1.5-Thinking-pro. These findings further underscore the effectiveness of structural representations in revealing subtle differences in model behavior.

\subsection{Additional Results on Best-of-N Decoding}
\label{appendix:sec3.4}
Table~\ref{appendix:tab:best_of_n} provides a detailed comparison of different Best-of-N decoding strategies on the MATH and LiveCodeBench (LCB) datasets using responses from two LLMs: DeepSeek-32B (DS-32) and Qwen-32B (QwQ-32). For the MATH benchmark, we evaluate on samples from the MATH500 and Level5 subsets that are not included in the training set. For LCB, we use LCB-v6 as the test set. In both cases, we ensure that the selected test samples are challenging—each sample is incorrectly answered at least twice across 10 runs. We set $N = 10$ and compare our proposed tree-based methods (Ours-Best and Ours-Vote) against several baselines:

\begin{itemize}
    \item Vote~\citep{wang2023selfconsistency}: Standard majority voting among $N$ outputs.
    \item Length-Best~\citep{wang2025thoughts}: Select the response with the fewest tokens.
    \item Length-Vote~\citep{wu2025more}: Majority voting after selecting the $k$ responses with reliable CoT length.
    \item ORM-Best~\citep{brown2024large}: Select the response with the highest outcome reward model score using Skywork-Reward-Gemma-2-27B-v0.2~\citep{liu2024skywork}.
    \item ORM-Vote~\citep{brown2024large}: Weighted Majority voting~\citep{lightman2023let} with the outcome reward model score.
    \item PRM-Best~\citep{zhang2025lessons}, which scores responses based on the product of step-level scores from a process reward model (i.e., Qwen2.5-Math-PRM-72B)
    \item PRM-Vote~\citep{zhang2025lessons}, Weighted Majority voting~\citep{lightman2023let} with the processing reward model score.
    \item Ours-Best: Select the response with the highest score assigned by our tree-based reasoning quality classifier mentioned in Section~\ref{sec3.3}.
    \item Ours-Vote: Weighted Majority voting with the score of our classifier.
\end{itemize}

Our method consistently outperforms traditional heuristics and reward model-based baselines, particularly in the MATH dataset, where precise multi-step reasoning is crucial. Notably, for DeepSeek-32B on MATH, our tree-based voting method achieves the highest accuracy at 82.47\%, significantly surpassing both Length-Best (56.70\%) and ORM-Best (60.82\%). Similar trends are observed for QwQ-32B, with our model showing competitive or superior performance. These results confirm that incorporating structural reasoning patterns via LCoT2Tree leads to a reliable output selection in complex reasoning tasks.

\section{Diagnostic Insight into Reasoning Behaviors \& Visualization Results}
\subsection{Insight into Error Behaviors}
\label{appendix:sec4.0}
\begin{figure*}[]
  \includegraphics[width=1.0\textwidth]{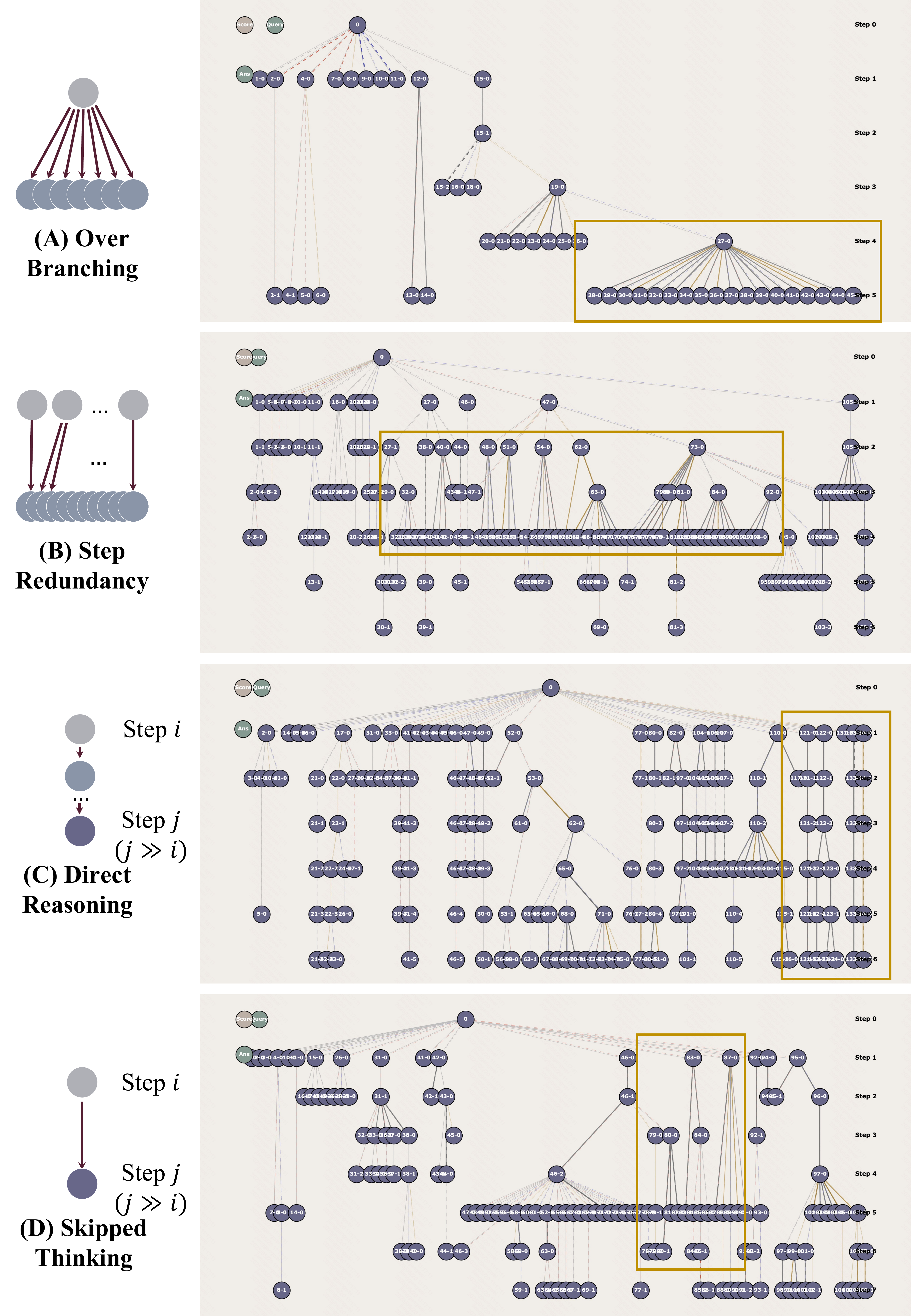}  
  \centering
  \caption{Visualization results of tree structure corresponding to different error patterns. The edge is labeled with the importance generated by GNNExplainer. The darker the color and the thicker the edge, the more important it is.  }
  \label{appendix:visual_error_case}
\end{figure*}

In this section, we present a detailed analysis of common error patterns found within reasoning trees. We use GNNExplainer~\citep{ying2019gnnexplainer}, a graph-based interpretability method, to identify which edges in a reasoning tree contribute most significantly to the model’s predictions. For each reasoning tree, GNNExplainer assigns an importance weight to every edge, reflecting its influence on the model’s output. These weights are normalized to the $[0, 1]$ range, and we visualize the tree by adjusting the edge thickness and color intensity according to these scores. The darker and thicker the edge, the more critical it is to the model's decision. Illustrative examples are shown in Figure~\ref{appendix:visual_error_case}.

Based on this analysis, we extract and categorize the most usual subgraphs associated with incorrect predictions into four primary error patterns. (A) Over Branching: excessive exploration or verification from a single node; (B) Step Redundancy: repetitive or unnecessary reasoning within the same step; (C) Direct Reasoning: abrupt transitions from one reasoning step to much deeper steps with minimal branching; (D) Skipped Thinking: leaping across multiple reasoning steps without proper intermediate logic.

These patterns are visualized in the left part of Figure~\ref{appendix:visual_error_case}, with real examples provided on the right. Notably, these findings reveal that both overly complex and overly simplistic reasoning paths can lead to incorrect outcome, underscoring the need for balanced, coherent, and well-structured reasoning in high-quality LLMs.

\subsection{Task-specific Reasoning Behaviors}
\label{appendix:sec4.1}
We have quantitatively demonstrated that LCoT2Tree effectively facilitates the separation of task-specific reasoning contents, as detailed in Section~\ref{subsec:4.2_task_specific}.
In this section, we leverage LCoT2Tree to pinpiont the disparate behaviors exhibited by the DeepSeek-32B model across various tasks. The key findings are summarized below:

For MATH (Figure~\ref{appendix:visual_ds32_math}), the reasoning trees typically display a diagonally descending structure, with progressively deeper steps achieved through repeated backtracking. This pattern reflects a structured, hierarchical problem-solving strategy. In the visualization, dashed lines—representing backtracking—are identified as key structural features that distinguish MATH from other tasks.

\begin{figure*}[]
  \includegraphics[width=1.05\textwidth]{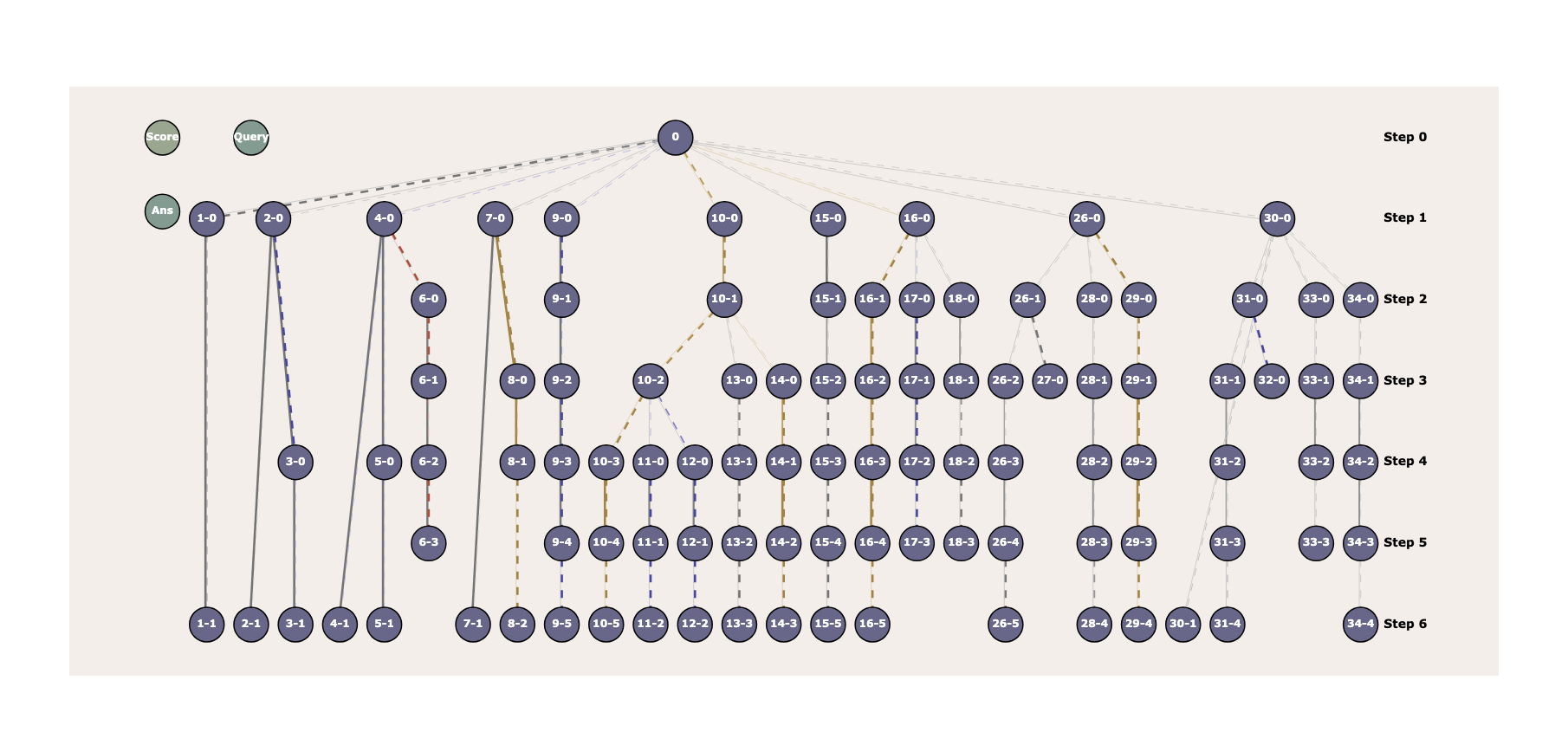}
  \centering
  \caption{Visualization results of tree structure of a response from \textbf{DeepSeek-32B} on \textbf{LiveCodeBench} dataset extracted using LCoT2Tree. The reasoning patterns tend to show broad, parallel branching, where many sibling nodes initiate independent linear thought without subsequent exploration or verification.}
  \label{appendix:visual_ds32_code}
\end{figure*}

For LiveCodeBench (Figure~\ref{appendix:visual_ds32_code}), the trees often exhibit broad, parallel branching, where many sibling nodes continue with independent linear thoughts that are rarely explored or verified further. This suggests a shallow, scattered reasoning style. Our visualization also reveals that these parallel branches contribute most significantly to classifying this task.

\begin{figure*}[]
  \includegraphics[width=1.05\textwidth]{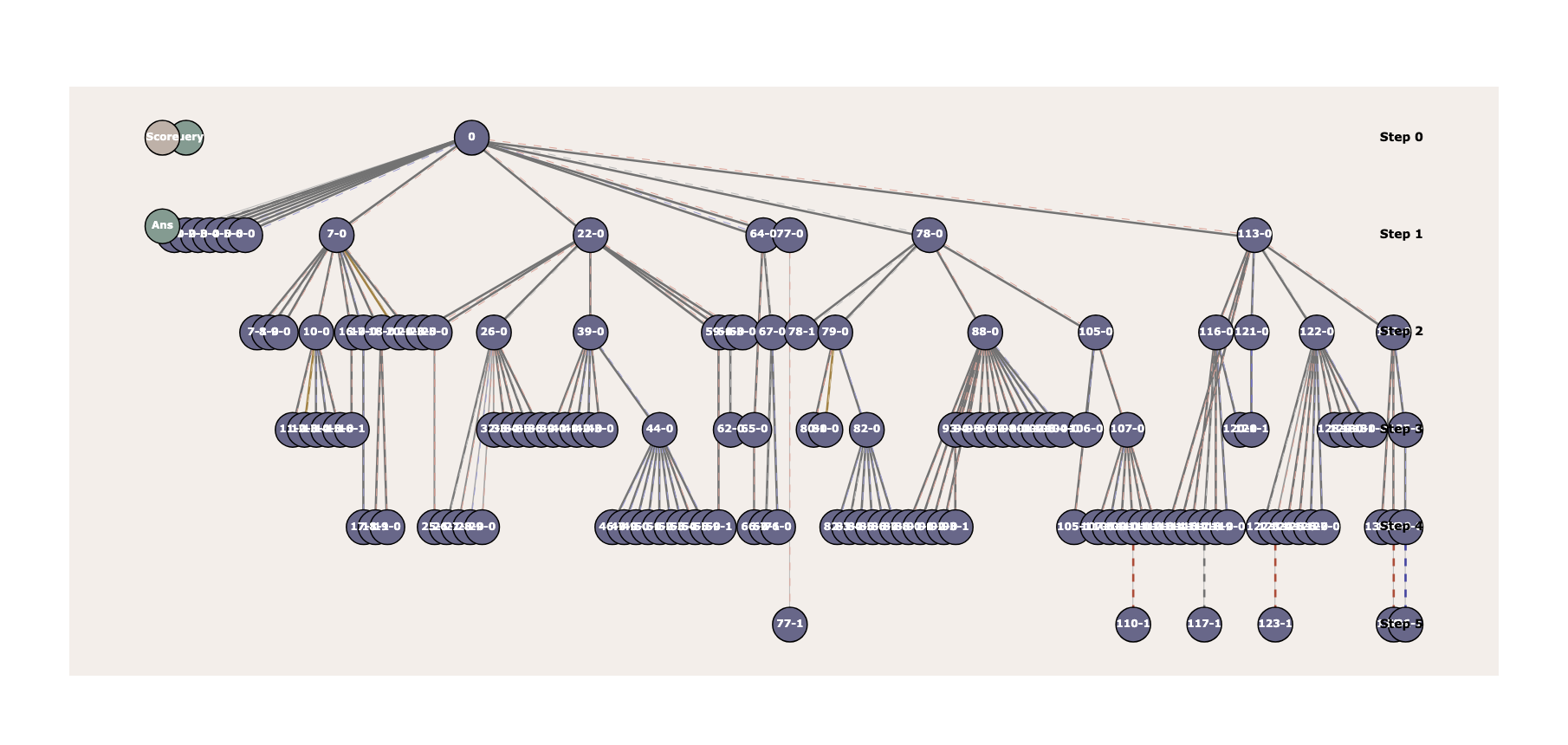}
  \centering
  \caption{Visualization results of tree structure of a response from \textbf{DeepSeek-32B} on \textbf{GPQA} dataset extracted using LCoT2Tree. The reasoning trees contain many high out-degree nodes, indicating that the model often revisits and elaborates on complex concepts.}
  \label{appendix:visual_ds32_gpqa}
\end{figure*}

For GPQA (Figure~\ref{appendix:visual_ds32_gpqa}), the reasoning trees contain numerous high out-degree nodes, indicating that the model frequently revisits and expands on specific concepts. This behavior suggests intensive cognitive effort and repeated clarification, reflecting the model's attempt to thoroughly understand difficult points—while also hinting at a lack of confidence in its reasoning.

\begin{figure*}[]
  \includegraphics[width=1.05\textwidth]{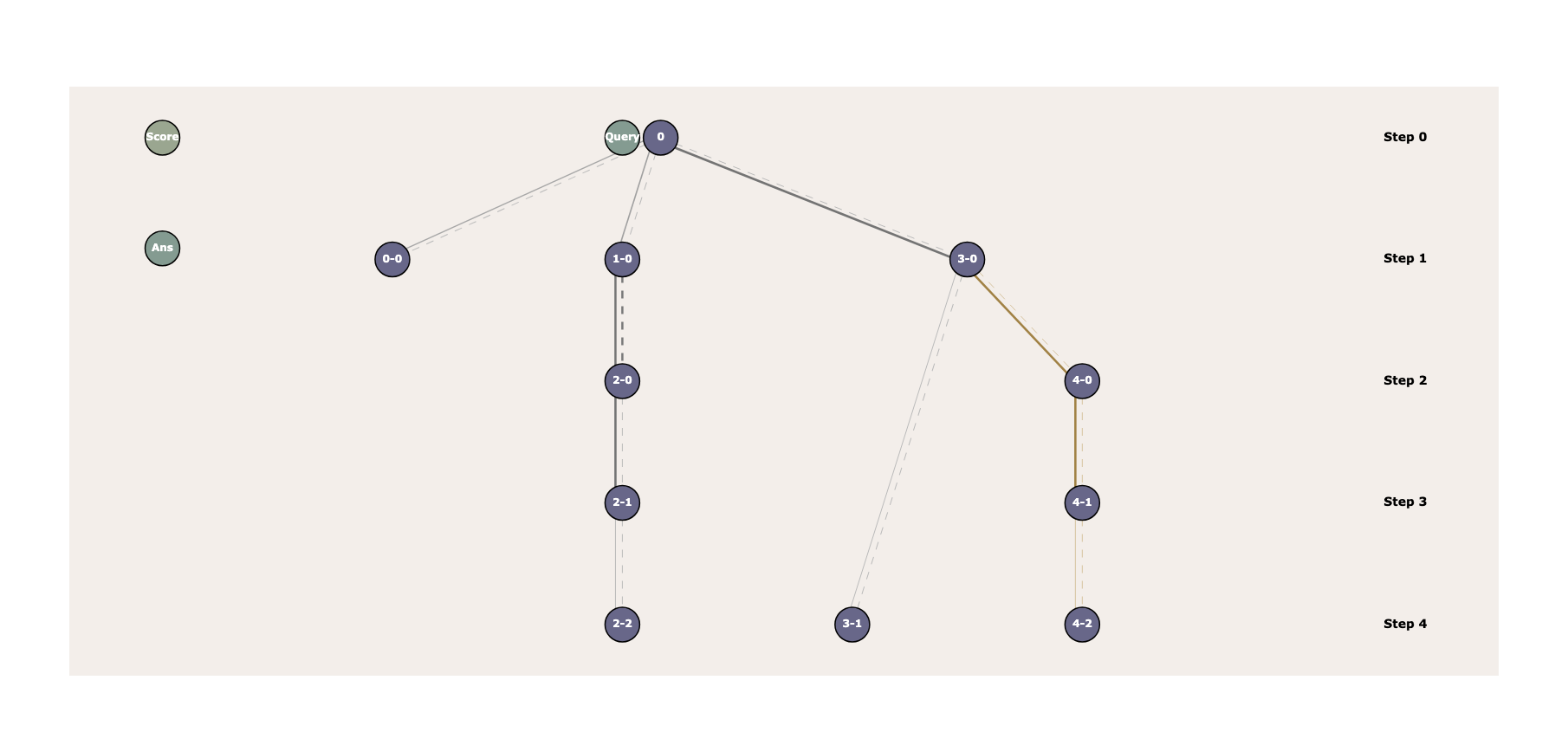}
  \centering
  \caption{Visualization results of tree structure of a response from \textbf{DeepSeek-32B} on \textbf{MMLU-Pro} dataset extracted using LCoT2Tree. The reasoning trees contain fewer nodes and minimal branching, indicating a more direct and deductive reasoning style with less exploration.}
  \label{appendix:visual_ds32_mmlu}
\end{figure*}

Finally, for MMLU-Pro(Figure~\ref{appendix:visual_ds32_mmlu}), the reasoning trees are relatively shallow, with fewer nodes and minimal branching. This suggests a more direct, deductive approach with limited exploration, which is consistent with the knowledge-intensive nature of MMLU-Pro questions rather than deeply compositional reasoning.

These observations highlight how LCoT2Tree provides fine-grained insights into the cognitive strategies employed by the model in diverse reasoning scenarios.

\subsection{Model-specific Reasoning Behaviors}
\label{appendix:sec4.2}
\begin{figure*}[]
  \centering
\includegraphics[width=1.05\textwidth]{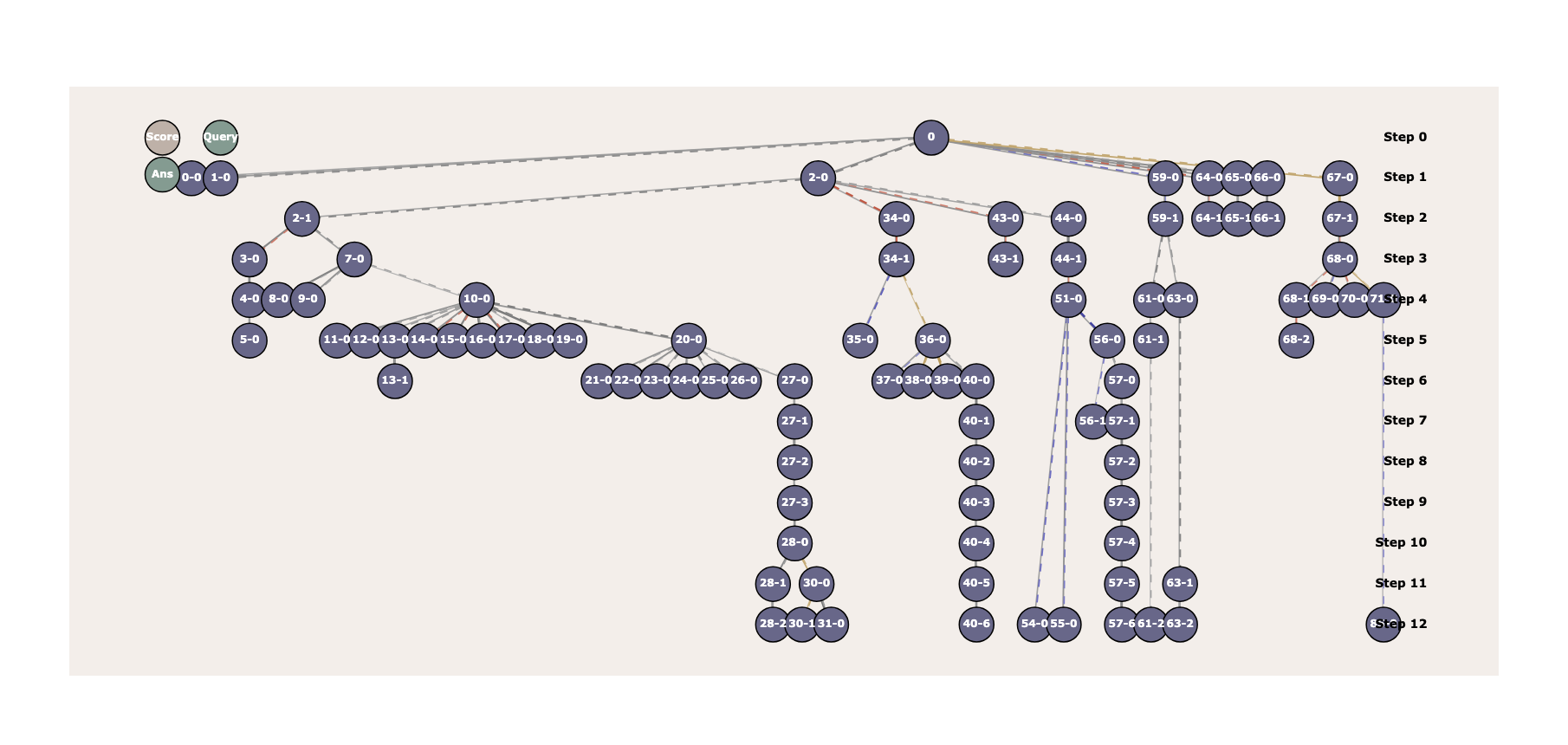}    
  \caption{Visualization results of tree structure of a response from \textbf{DeepSeek-32B} on \textbf{MATH} dataset extracted using LCoT2Tree. The reasoning trees exhibit a downward-sloping hierarchical structure, with progressively deeper steps achieved through repeated backtracking. }
  \label{appendix:visual_ds32_math}
\end{figure*}

We provide a detailed comparison of how different LLMs approach the same task by visualizing and analyzing their reasoning trees on the MATH dataset. Focusing on DeepSeek-32B as a reference point, we summarize several key observations:

DeepSeek-32B (DS-32; Figure~\ref{appendix:visual_ds32_math}) typically produces reasoning trees with a diagonally descending structure, with depth increasing progressively through backtracking. This reflects a structured, step-by-step problem-solving reasoning process.

\begin{figure*}[]
  \includegraphics[width=1.05\textwidth]{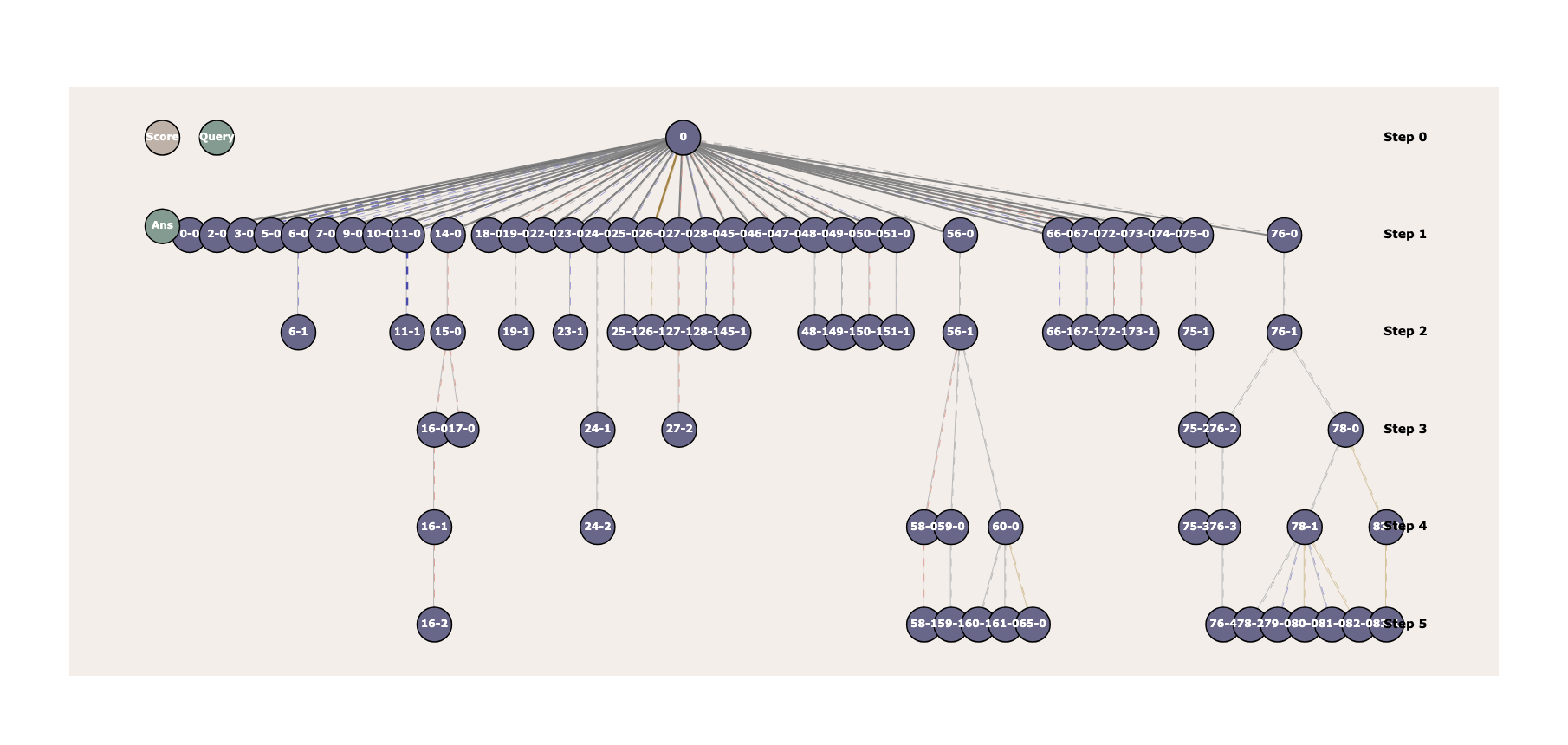}
  \centering
  \caption{Visualization results of tree structure of a response from \textbf{DeepSeek-R1} on \textbf{MATH} dataset extracted using LCoT2Tree. It exhibits similar behavior to DS-32, but with an important distinction: it tends to truncate detailed exploration earlier and backtrack to beginning steps more quickly to optimize its reasoning path. }
  \label{appendix:visual_dsr1_math}
\end{figure*}

DeepSeek-R1 (Figure~\ref{appendix:visual_dsr1_math}) exhibits similar structural characteristics to DS-32, but with a notable difference: it tends to terminate detailed exploration earlier and backtrack more quickly to beginning steps. This indicates a more aggressive pruning strategy to streamline the reasoning path. In visualizations, connections between Step 0 and Step 1 serve as critical features distinguishing DeepSeek-R1's behavior.

\begin{figure*}[]
  \includegraphics[width=1.05\textwidth]{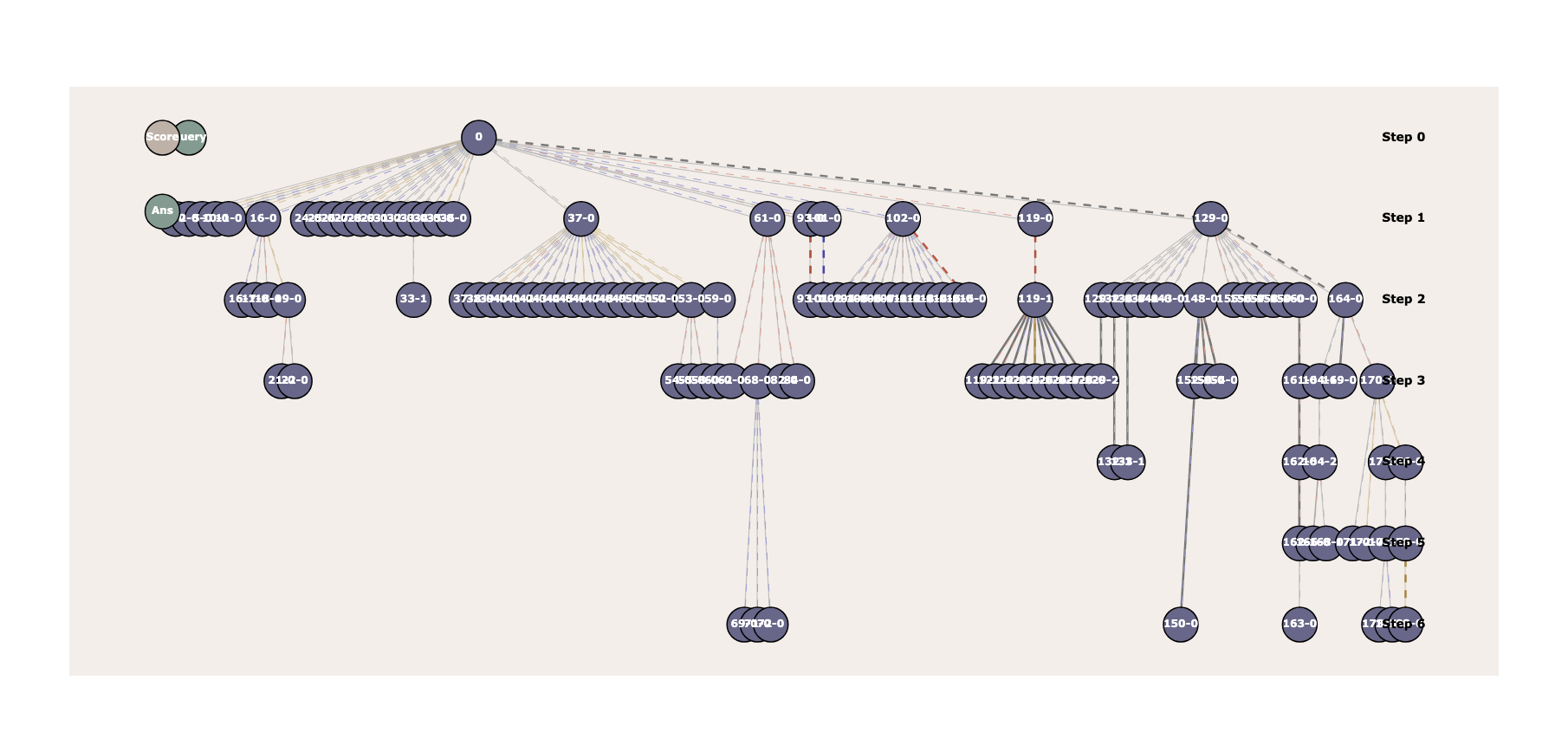}
  \centering
  \caption{Visualization results of tree structure of a response from \textbf{QwQ-32B} on \textbf{MATH} dataset extracted using LCoT2Tree. QwQ-32B mirrors the behavior of DeepSeek-32B to some extent, but differs in how it allocates attention in the latter stages of reasoning.}
  \label{appendix:visual_qwq_math}
\end{figure*}

QwQ-32B (Figure~\ref{appendix:visual_qwq_math}) also mirrors the behavior of DS-32 to some extent but differs in the latter stages. Unlike DS-32, which often rushes toward the final answer, QwQ-32B continues to invest cognitive effort into deeper exploration. In the visualization, expanded right subtrees often emerge as defining characteristics of QwQ-32B’s reasoning tree.

\begin{figure*}[]
  \includegraphics[width=1.05\textwidth]{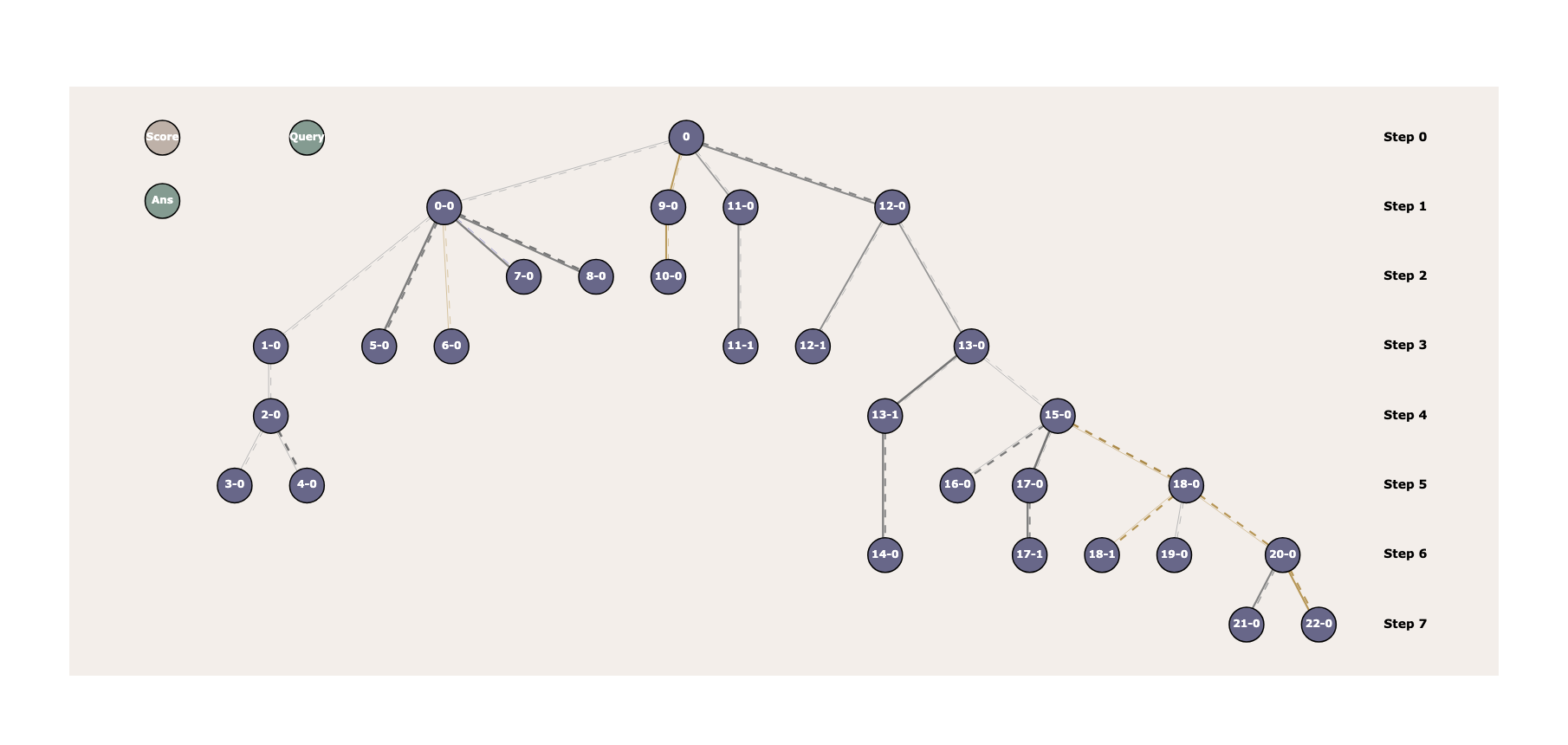}
  \centering
  \caption{Visualization results of tree structure of a response from \textbf{Seed-1.5-Thinking-pro} on \textbf{MATH} dataset extracted using LCoT2Tree. The reasoning trees contain fewer nodes and branches, forming simpler structures.}
  \label{appendix:visual_seed_math}
\end{figure*}

\begin{figure*}[h]
\centering
\includegraphics[width=1.05\textwidth]{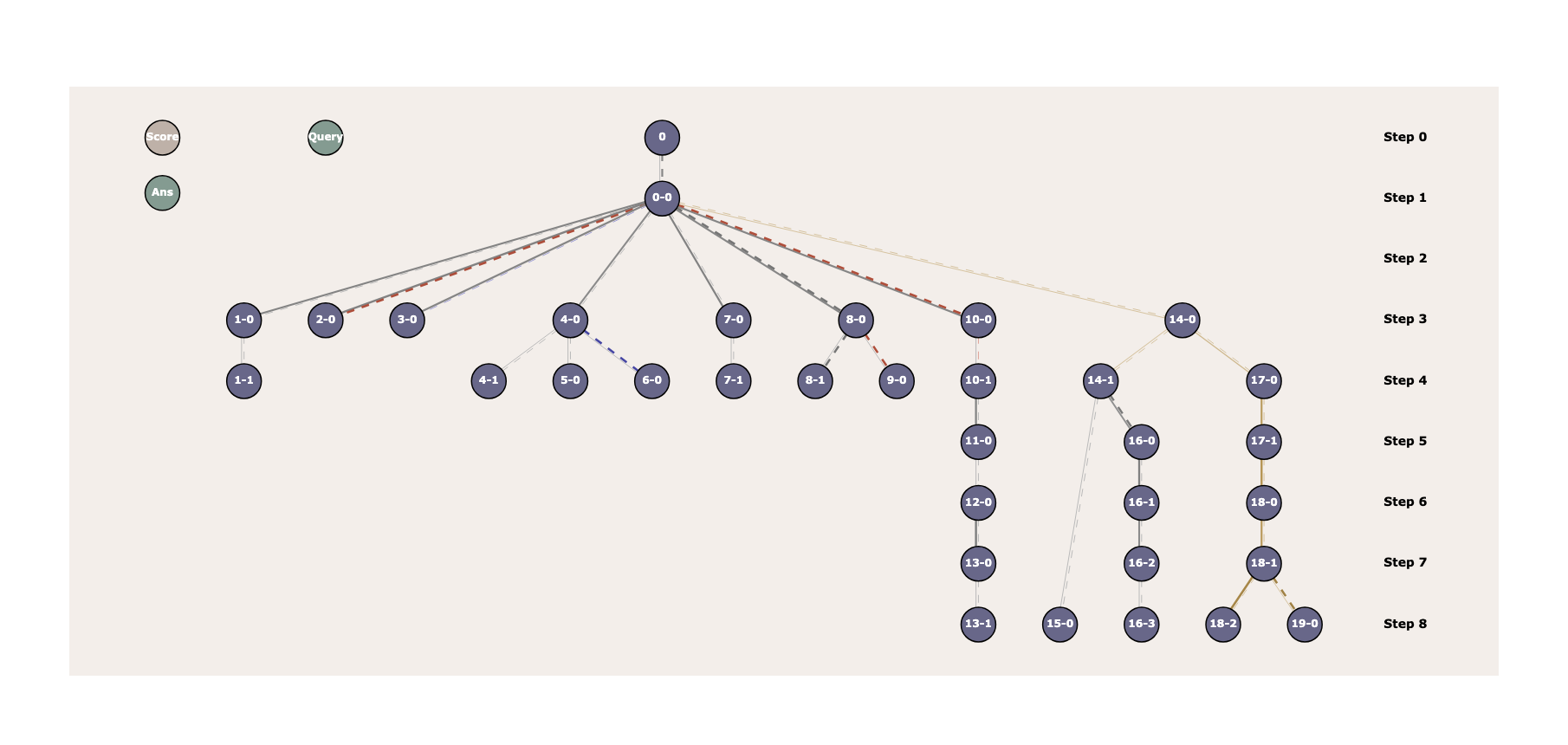}
  \caption{Visualization results of tree structure of a response from \textbf{Grok-3-mini-beta} on \textbf{MATH} dataset extracted using LCoT2Tree. The reasoning trees contain fewer nodes and branches, forming simpler structures.}
\label{appendix:visual_grok_math}
\end{figure*}

In contrast, Seed-1.5-Thinking-pro (Figure~\ref{appendix:visual_seed_math}) and Grok-3-mini-beta (Figure~\ref{appendix:visual_grok_math}) follow a markedly different reasoning strategy. They exhibit fewer thought transitions during reasoning. As a result, their trees contain fewer nodes and branches, forming simpler structures. This suggests a straightforward problem-solving style with limited iterative refinement.

These insights reinforce that LCoT2Tree not only captures reasoning structure at the task level, but also reveals distinctive behavioral patterns across model families.

\begin{figure*}[b]
\centering
\begin{tcolorbox}[colback=white,colframe=black,title=Step1 Prompt in LCoT2Tree tool to extract reasoning sketch from LCoT]
Analyze the following reasoning text and  extract a strictly ordered, atomic sequence of key reasoning steps. Focus on extracting the validated, logically essential progression of thoughts while excluding backtracking, rechecks, or redundant details.  
\newline

Reasoning text: 

<reasoning\_text>

\textbf{\{\{text\}\}}

</reasoning\_text>
\newline

Please read the entire text carefully and generate by following these rules:

1. Find the key steps and the logical flow of reasoning. 

2. Each step must represent a single, indivisible logical action that directly advances the reasoning.

3. Determine the correct version of the step, ignoring redundant information. A correct step should be able to push the reasoning logic forward and have no errors in itself.

4. Do not skip steps. Do not merge steps. Use the original phrasing where possible.

5. Do not include verification steps unless it introduces new constraints.

6. Organize the steps into a coherent sequence of key reasoning steps and number it sequentially (1., 2., 3., ...). 

7. Maintain strict output format.
\newline

Output format:

<reasoning\_process>

Step 1. concise statement: Detail step

Step 2. concise statement: Detail step

Step 3. concise statement: Detail step

</reasoning\_process>
\newline

Please list the key reasoning steps of the provided text. 
\end{tcolorbox}
\caption{The content of Step1 Prompt in LCoT2Tree tool to extract reasoning sketch from LCoT.}
\label{appendix:prompt1}
\end{figure*}

\begin{figure*}[]
\centering
\begin{tcolorbox}[colback=white,colframe=black,title=Step3 Prompt in LCoT2Tree tool to assign reasoning step to each thought.]
Your task is to match each reasoning thought from List B to corresponding step number(s) in the List A. Follow the following process:
\newline

1. First understand List B:

   - For each thought in List B, identify if it describes some specific calculation processes (mathematical operation, logical transformation, or data manipulation)

   - Ignore the describation that only state conclusions, concepts without showing the actual processing detail
\newline

2. Then math to List A:
   
   - For each thought from List B, find all steps in List A that:

     * Show the same underlying calculation (even with different numbers/words)
     
     * Represent the partial or same reasoning process

   - Ignore superficial wording differences - focus on logical equivalence
\newline

3. Output requirements:

   - Return ALL plausible matches where computational processes align
   
   - Never return empty arrays (except for thought B0 if needed)
   
   - Multiple matches are encouraged when justified
   
   - Maintain strict JSON format
\newline

Input:

- List A (Detailed Steps): 

<list\_a>

\textbf{\{\{reasoning\_step\}\}}

</list\_a>

- List B (Reasoning Thoughts): 

<list\_b>

\textbf{\{\{thoughts\}\}}

</list\_b>
\newline

Output Format (strict JSON):

```json

\{

  "B0": ["A1"],
  
  "B1": ["A3"],
  
  "B2": ["A1", "A4"],
  
  ...
  
\}'''
\newline

Please match the reasoning thoughts in List B to step in the List A.

\end{tcolorbox}
\caption{The content of Step3 Prompt in LCoT2Tree tool to assign reasoning step to each thought.}
\label{appendix:prompt2}
\end{figure*}

\begin{figure*}[]
\centering
\begin{tcolorbox}[colback=white,colframe=black,title=Step4 Prompt in LCoT2Tree tool to assign function to each thought.]
Your task is to classify Text2's purpose relative to Text1 using these categories:
\newline

Categories:

1. Continuous Logic - Direct continuation/extension of Text1's reasoning flow

2. Exploration - Introduces parallel/unrelated concepts from Text1, alternative reasoning paths, or new topics

3. Backtracking - Revises, corrects, or adjusts previous step

4. Validation - Provides supporting evidence, logical justification, or examples for Text1's claims
\newline

Input:
\textbf{\{\{}

\textbf{  "Text1": "{TEXT1}",}

\textbf{  "Text2": "{TEXT2}"}

\textbf{\}\}}
\newline

Output Format:

Return only JSON format ```json\{"Category": "Name of Category"\}'''
\end{tcolorbox}
\caption{The content of Step4 Prompt in LCoT2Tree tool to assign function to each thought.}
\label{appendix:prompt3}
\end{figure*}

\end{document}